\documentclass[11pt]{article}

\usepackage[final]{acl}

\usepackage{times}
\usepackage{latexsym}
\usepackage{booktabs}
\usepackage{multirow}
\usepackage{amsfonts}
\usepackage{amsmath}
\usepackage{tabularx}
\usepackage{kotex}
\usepackage{xcolor} 
\usepackage[table]{xcolor}
\usepackage{fvextra}

\usepackage[T1]{fontenc}
\usepackage[utf8]{inputenc}

\usepackage{inconsolata}

\usepackage{graphicx}

\title{Understanding LLM Behavior in Multi-Target Cross-Lingual Summarization}

\author{
\textbf{Sangwon Ryu$^{1}$, Yihong Liu$^{3,4}$, Mingyang Wang$^{3,4}$,} \\
\textbf{Yunsu Kim$^{5}$, Jungseul Ok$^{1,2}$, Gary Geunbae Lee$^{1,2}$, Hinrich Schütze$^{3,4}$} \\
$^{1}$GSAI, POSTECH \quad
$^{2}$CSE, POSTECH \\
$^{3}$LMU Munich \quad
$^{4}$MCML \quad
$^{5}$LILT\\
\texttt{ryusangwon@postech.ac.kr} \\
}

\begin{document}
\maketitle

\begin{abstract}

Multi-target cross-lingual text summarization (MTXLS), which summarizes a source document into multiple target languages, is increasingly important as users consume content in diverse languages, but remains underexplored.
To address this gap, we introduce \underline{m}ulti-target cross-lingual \underline{e}lement-\underline{a}ware (MEA), a new MTXLS benchmark covering 24 target languages. We benchmark end-to-end and pipeline approaches across various LLMs and show that MTXLS performance still substantially lags behind English monolingual summarization.
To better understand MTXLS in LLMs, we propose a layer-wise analysis framework for investigating how LLMs internally perform MTXLS. 
Our analyses suggest that translation and summarization behaviors emerge jointly within later layers rather than as distinctly decomposed stages. Most task-relevant processing occurs within these layers, and errors also tend to arise at similar depths.
Motivated by these findings, we introduce an inference-time activation steering method that leverages hidden representations from English summarization to guide MTXLS generation. Experiments show that our method consistently improves MTXLS quality across target languages.

\end{abstract}
\section{Introduction}

Text summarization plays an important role in helping users efficiently access large amounts of information. Yet much globally accessible information is written in English, while users worldwide consume content in diverse languages. This makes multi-target cross-lingual summarization (MTXLS), which summarizes a source document into multiple target languages, increasingly important for multilingual information access.

Despite its practical importance, MTXLS remains underexplored, with only recent work beginning to formalize the task \cite{pernes-etal-2024-multi}. Although LLMs show strong high-resource translation and English summarization capabilities individually~\cite{zhu-etal-2024-multilingual, goyal2022news, pu2023summarization, zhang-etal-2024-benchmarking, ryu24_interspeech, ryu2025adaptive}, it remains unclear how effectively they perform MTXLS and whether end-to-end (E2E) or pipeline approaches that separately perform summarization and translation are more suitable.

To address these challenges, we introduce \underline{m}ulti-target cross-lingual \underline{e}lement-\underline{a}ware (MEA), a new benchmark covering 24 target languages based on an element-aware test set~\cite{wang-etal-2023-element} designed to preserve key information consistently across languages. Using MEA, we benchmark E2E and pipeline approaches across various LLMs and find that MTXLS performance still lags behind English summarization. We also observe substantial inconsistencies across methods and target languages. Even with the same backbone model and source document, generated summaries can differ substantially across generation strategies and exhibit language-dependent hallucination and omission patterns.

To better understand these limitations, we propose a layer-wise analysis framework that tracks the emergence of translation and summarization across model layers. 
Our analyses suggest that these behaviors tend to emerge jointly within later layers, rather than as distinct stages.
Moreover, errors tend to emerge around the same layers where translation and summarization behaviors arise. We further find that these layer-wise patterns remain largely consistent across target languages within the same model, while different model families exhibit distinct trends.

Motivated by these findings, we introduce an inference-time activation steering method that leverages hidden representations from English summarization to guide generation. Experiments show that our method consistently improves generation quality across target languages.
Overall, this paper makes four main contributions:

\begin{itemize}
    \item We introduce MEA, a new MTXLS benchmark covering 24 target languages.
    
    \item We systematically benchmark E2E and pipeline approaches for MTXLS across various LLMs.
    
    \item We propose a layer-wise analysis framework that localizes translation- and summarization-related behaviors, as well as hallucinations and omissions.

    \item We present an activation steering method that leverages hidden representations from English summarization to improve MTXLS across languages.
\end{itemize}
\section{Related Work}

% \paragraph{Cross-lingual Summarization} 

Cross-lingual summarization (XLS), which summarizes a source document into a target language, has been widely studied due to its practical relevance. Early approaches primarily relied on pipeline methods that decouple translation and summarization~\cite{leuski-hindi-2003, orasan-chiorean-2008-evaluation, wan-etal-2010-cross, wan-2011-using,Zhang-trans-2016, ouyang-etal-2019-robust}, while subsequent works explored building E2E models for specific language pairs~\cite{zhu-etal-2019-ncls, ladhak-etal-2020-wikilingua, bai-etal-2021-cross, wang-etal-2022-clidsum, wang-etal-2022-survey, feng-etal-2022-msamsum, chen-etal-2023-revisiting,  bhattacharjee-etal-2023-crosssum}. More recently, studies have leveraged LLMs to perform XLS in a unified manner~\cite{wang-etal-2023-zero, park-etal-2024-low, huot-etal-2024-mplan, wang-etal-2025-empirical}, eliminating the need for separate models for each language pair. In particular, \citet{pernes-etal-2024-multi} introduced MTXLS and showed that LLMs often fail to produce coherent summaries consistently across different languages. However, prior work remains limited by the lack of datasets with broad and consistent language coverage, making systematic evaluation difficult in multi-target settings. Moreover, the internal processing behaviors of LLMs in MTXLS remain largely unexplored, particularly regarding how they vary across target languages.

% \paragraph{Mechanistic Interpretability}

\begin{table*}[t]
\centering
\small
\setlength{\tabcolsep}{5pt}
\resizebox{\textwidth}{!}{
\begin{tabular}{l|c|cccc|cccc|cccc}
\toprule
\multirow{2}{*}{\textbf{Model}} 
& \multirow{2}{*}{\textbf{Method}}
& \multicolumn{4}{c|}{\textbf{High}}
& \multicolumn{4}{c|}{\textbf{Mid}}
& \multicolumn{4}{c}{\textbf{Low}} \\
\cmidrule(lr){3-6} \cmidrule(lr){7-10} \cmidrule(lr){11-14}
& 
& \textbf{Coh.} & \textbf{Cons.} & \textbf{Flu.} & \textbf{Rel.}
& \textbf{Coh.} & \textbf{Cons.} & \textbf{Flu.} & \textbf{Rel.}
& \textbf{Coh.} & \textbf{Cons.} & \textbf{Flu.} & \textbf{Rel.} \\
\midrule

Qwen3.5-2B & S-T
& \textbf{3.84} & 3.71 & \textbf{4.27} & 3.96
& \textbf{3.49} & 3.28 & 3.37 & \textbf{3.59}
& 1.31 & 1.55 & 1.61 & 1.56 \\
& T-S
& 3.76 & \textbf{3.79} & 4.23 & \textbf{3.98}
& 3.25 & \textbf{3.29} & \textbf{3.65} & 3.47
& \textbf{1.46} & \textbf{1.67} & \textbf{1.88} & \textbf{1.67} \\
& E2E
& 3.70 & 3.65 & 4.14 & 3.87
& 3.07 & 3.09 & 3.13 & 3.23
& 1.08 & 1.26 & 1.43 & 1.19 \\
\midrule

Tiny-Aya-Global & S-T
& 4.05 & 4.09 & 4.74 & 4.29
& 3.98 & 3.95 & 4.35 & 4.12
& \textbf{3.01} & 3.20 & 3.59 & \textbf{3.10} \\
& T-S
& 4.04 & 4.14 & 4.68 & 4.25
& 3.92 & 4.03 & 4.37 & 4.09
& 1.57 & 2.02 & 2.69 & 1.62 \\
& E2E
& \textbf{4.12} & \textbf{4.25} & \textbf{4.75} & \textbf{4.37}
& \textbf{4.02} & \textbf{4.12} & \textbf{4.49} & \textbf{4.20}
& 3.00 & \textbf{3.25} & \textbf{3.61} & 3.09 \\
\midrule

Qwen3.5-9B & S-T
& 4.23 & 4.50 & \textbf{4.87} & 4.62
& 4.07 & 4.23 & 4.72 & 4.40
& 3.94 & 4.02 & 4.44 & 4.10 \\
& T-S
& 4.22 & 4.51 & 4.86 & 4.60
& 4.07 & 4.23 & 4.74 & 4.38
& \textbf{4.00} & 4.01 & \textbf{4.54} & 4.11 \\
& E2E
& \textbf{4.25} & \textbf{4.57} & \textbf{4.87} & \textbf{4.65}
& \textbf{4.09} & \textbf{4.32} & \textbf{4.75} & \textbf{4.45}
& 3.93 & \textbf{4.08} & 4.41 & \textbf{4.15} \\
\midrule

gpt-oss-20b & S-T
& 4.12 & 4.32 & 4.80 & \textbf{4.43}
& 3.96 & \textbf{4.06} & 4.50 & 4.17
& \textbf{3.41} & \textbf{3.50} & 3.46 & \textbf{3.51} \\
& T-S
& 3.72 & 3.89 & 4.80 & 3.88
& 3.67 & 3.71 & 4.52 & 3.68
& 3.32 & 3.24 & \textbf{4.03} & 3.48 \\
& E2E
& \textbf{4.14} & \textbf{4.33} & \textbf{4.82} & \textbf{4.43}
& \textbf{4.00} & \textbf{4.06} & \textbf{4.56} & \textbf{4.18}
& 3.15 & 3.28 & 3.27 & 3.23 \\

\bottomrule
\end{tabular}
}
\caption{MTXLS performance across models, resource levels, and methods. S-T, T-S, and E2E denote summarize-then-translate, translate-then-summarize, and end-to-end cross-lingual summarization, respectively. E2E generally achieves stronger performance for larger or multilingual-oriented models. \textbf{Bold} indicates the best scores within each model and resource group.}
\label{tab: geval_dimensions}
\end{table*}

\section{Benchmarking MTXLS in LLMs}
\label{section: benchmarking}

Prior work has often motivated E2E XLS by arguing that pipeline approaches suffer from error propagation between translation and summarization stages~\cite{leuski-hindi-2003, orasan-chiorean-2008-evaluation, wan-etal-2010-cross, wan-2011-using, Zhang-trans-2016, ouyang-etal-2019-robust}. However, much of this evidence comes from supervised encoder-decoder settings, leaving it unclear whether the same conclusion holds for decoder-only models. We therefore systematically compare E2E and pipeline methods under zero-shot LLMs.

\subsection{Multi-target cross-lingual element-aware dataset (MEA)}
\label{section: mea}

To enable systematic evaluation of LLM-based MTXLS across E2E and pipeline settings, we construct MEA, a new benchmark based on the CNN/DailyMail element-aware dataset~\cite{wang-etal-2023-element}, a high-quality English summarization test set curated by experts to ensure preservation of salient information elements. We extend this dataset to the multi-target cross-lingual setting by translating all examples into 24 target languages spanning diverse language families and resource levels using Google Translate (v3 Translate LLM)~\footnote{\url{https://cloud.google.com/translate}}: Chinese (zh), Spanish (es), French (fr), German (de), Japanese (ja), Russian (ru), Portuguese (pt), Korean (ko), Italian (it), Dutch (nl), Polish (pl), Turkish (tr), Vietnamese (vi), Arabic (ar), Ukrainian (uk), Hindi (hi), Greek (el), Thai (th), Hungarian (hu), Persian (fa), Bengali (bn), Burmese (my), Khmer (km), and Georgian (ka).
To ensure translation quality, we further employ GPT-4o-mini~\cite{hurst2024gpt} to validate each instance along four dimensions: \textit{hallucination}, \textit{language mismatch}, \textit{missing entity}, and \textit{naturalness}. Instances with identified issues are regenerated to improve reliability.
In total, we translate all 200 examples from the CNN/DailyMail element-aware test set into 24 languages, resulting in a 4.8K-instance benchmark. Additional details are provided in Appendix~\ref{appendix: MEA_construction}.

\subsection{Experimental setup}

\paragraph{Models.}
We evaluate a diverse set of language models, including Qwen3.5 (2B and 9B)~\cite{qwen3.5}, Tiny-Aya-Global (3.35B)~\cite{salamanca2026tinyaya}, and gpt-oss-20b~\cite{agarwal2025gpt}, to enable comprehensive comparison.

\paragraph{Metrics.}
To assess MTXLS performance, we adopt XLS-adapted variants of G-Eval~\cite{liu-etal-2023-g}, \texttt{multilingual ROUGE}~\cite{lin-2004-rouge, hasan-etal-2021-xl}, and BERTScore~\cite{bert-score} with \texttt{bert-base-multilingual-cased}~\cite{devlin-etal-2019-bert}. 
Since \texttt{multilingual-rouge} and BERTScore support a limited set of target languages, we primarily rely on G-Eval as our main evaluation metric. Details of the evaluation setup for each language are provided in Appendix~\ref{appendix: evaluation_details}.

\subsection{End-to-End vs. Pipeline MTXLS}

We first evaluate MTXLS in current LLMs by comparing E2E generation with two pipeline approaches: summarize-then-translate (S-T) and translate-then-summarize (T-S). Prior work has often motivated E2E XLS by suggesting that pipeline approaches suffer from error propagation between summarization and translation stages~\cite{park-etal-2024-low, wang-etal-2025-empirical}. However, recent advances in LLMs, including open-source models, have substantially improved zero-shot translation and summarization capabilities, motivating a renewed comparison between E2E and pipeline approaches.

Following the language resource classification of \citet{joshi-etal-2020-state}, we categorize languages into high-resource (4–5), mid-resource (3), and low-resource (1–2) groups (Table~\ref{tab: language_resource_level}). 
Table~\ref{tab: geval_dimensions} summarizes results by resource level, while Appendix~\ref{appendix: full_results} provides the full results, including English summarization.
Across all settings, XLS consistently underperforms English summarization, highlighting that \textbf{current LLMs still struggle with cross-lingual generation}.

For smaller models, such as 2B-scale LLMs, pipeline approaches generally outperform E2E generation. However, E2E methods consistently achieve stronger overall performance as models either become larger, such as Qwen3.5-9B, or more multilingual-oriented, such as Tiny-Aya-Global. In particular, the T-S pipeline often suffers substantial degradation, suggesting that errors introduced during document-level translation propagate to downstream summarization.
% ; qualitative examples of such error propagation are provided in the Appendix~\ref{appendix: error_propagation_examples}.
Overall, despite requiring only a single generation pass, \textbf{zero-shot E2E generation performs strongly in most settings, highlighting its potential as an effective and scalable alternative for XLS.}

\subsection{Inconsistency across methods}

Interestingly, different MTXLS methods often generate markedly different summaries even when using the same backbone model and source document. As shown in Table~\ref{tab: inconsistency-across-methods}, the summaries differ not only in wording, but also in factual certainty and narrative framing. For example, while the Gold summary describes the event as an ``allegation,'' T-S adopts a more assertive framing (``sanctions were imposed due to''). Likewise, S-T generates a more accusatory description (``officially accused'') than gold summary. These observations suggest that \textbf{different generation strategies can influence how models prioritize and frame information, leading to meaningful semantic divergence across summaries}.

\subsection{Inconsistency across target languages}

We further observe that the same model can exhibit different error patterns across target languages. For example, the primary focus of the summary shift across languages, while low-resource languages often exhibit more severe hallucinations and factual inconsistencies. 
As shown in Table~\ref{tab: inconsistency-across-languages}, the model produces different outputs across German, Hungarian, and Turkish, all of which use the Latin script.
In the high-resource language German, the model correctly generates ``Scottish Cup semifinal.'' In contrast, the Hungarian summary incorrectly produces ``Scottish Cup Final,'' while the Turkish summary generates ``Champions Cup semifinal.'' These findings suggest that \textbf{LLMs remain prone to language-dependent instability and hallucinations in MTXLS, highlighting the need for further research toward more reliable MTXLS systems}.

\begin{figure*}[ht]
\centering
\includegraphics[width=\textwidth]{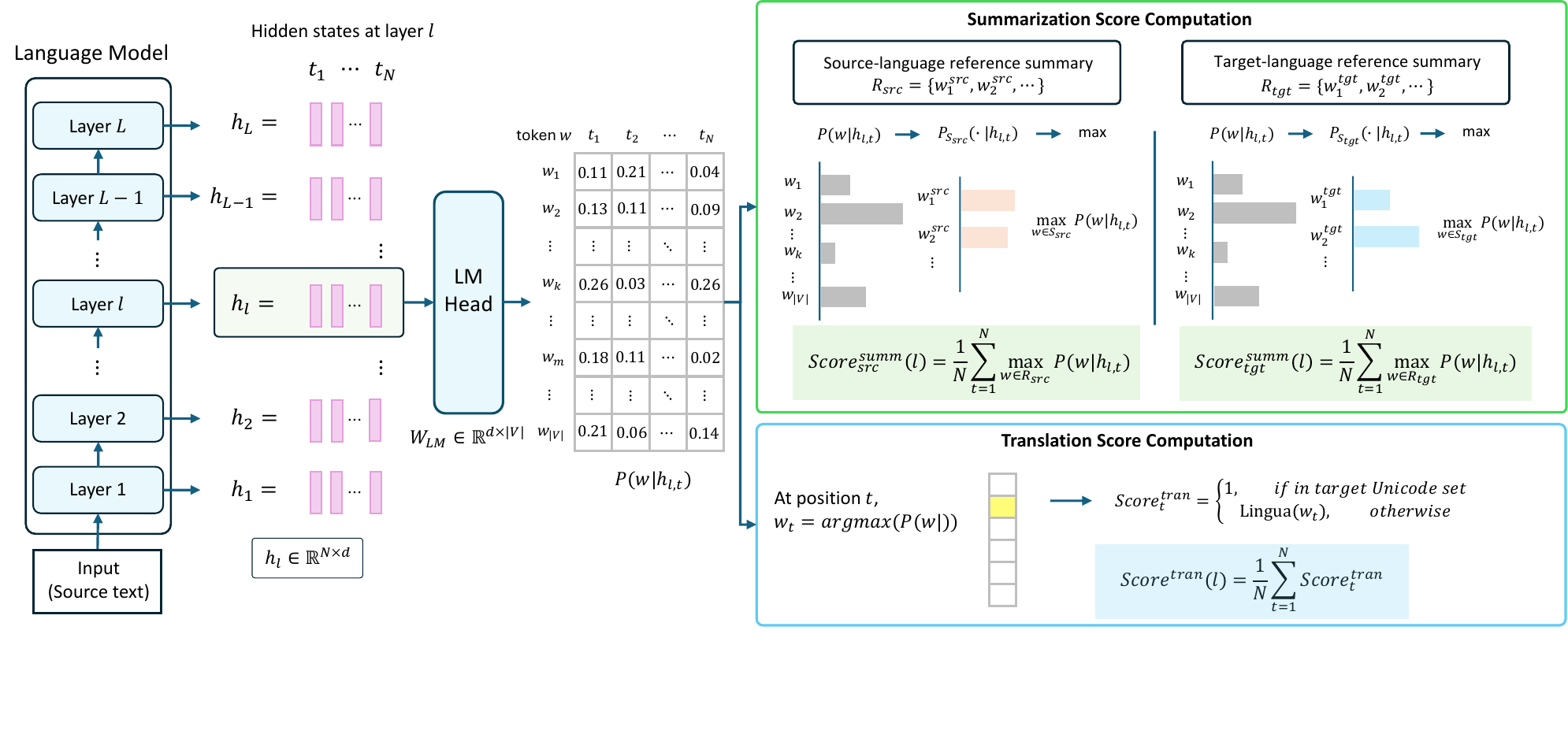}
\caption{Illustration of our layer-wise MTXLS analysis framework. We compute translation and summarization scores from intermediate hidden states to examine how these behaviors emerge across layers.}
\label{fig: layer-wise-analysis-method}
\end{figure*}

% \begin{table}[t]
% \centering
% \scriptsize
% \setlength{\tabcolsep}{3pt}

% \begin{tabular}{p{0.8cm}|p{5.3cm}}
% \toprule
% \textbf{Method} & \textbf{Summary Excerpt} \\
% \midrule

% Gold &
% ``위키백과 온라인 약력을 조작했다는 \textit{혐의}를 받은 ...
% 혐의가 거짓이라고 강조했다.'' \\

% \midrule

% S-T &
% ``위키백과 프로필을 조작했다는 \textit{혐의로 그를 공식적으로 고발}한 인물로 드러났습니다 ...
% `처음부터 끝까지 거짓'이라고 격분하며 부인했고 ...'' \\

% \midrule

% T-S &
% ``위키백과 전기 조작을 \textit{이유로 차단 조치}를 내린 사실이 어제 드러났습니다 ...
% 이를 완전히 거짓이라고 강력히 부인했습니다.'' \\

% \midrule

% E2E &
% ``위키백과 프로필이 조작되었다는 \textit{의혹이 제기된 사건}에서 ...
% `사실과 전혀 다르다'며 강력히 부인했습니다.'' \\

% \bottomrule
% \end{tabular}

\begin{table}[t]
\centering
\scriptsize
\setlength{\tabcolsep}{3pt}

\begin{tabular}{p{0.9cm}|p{5.5cm}}
\toprule
\textbf{Method} & \textbf{Summary Excerpt} \\
\midrule

Gold &
``위키백과 온라인 약력을 조작했다는 \textit{혐의}를 받은 ...
혐의가 거짓이라고 강조했다.''

{\tiny \textcolor{gray}{
(“He was accused of manipulating a Wikipedia online profile ...
and emphasized that the allegation was false.”)
}} \\

\midrule

T-S &
``위키백과 전기 조작을 \textit{이유로 차단 조치}를 내린 사실이 어제 드러났습니다 ...
이를 완전히 거짓이라고 강력히 부인했습니다.''

{\tiny \textcolor{gray}{
(“It was revealed yesterday that sanctions were imposed due to manipulation of a Wikipedia biography ...
and he strongly denied it as completely false.”)
}} \\

\midrule

S-T &
``위키백과 프로필을 조작했다는 \textit{혐의로 그를 공식적으로 고발}한 인물로 드러났습니다 ...
`처음부터 끝까지 거짓'이라고 격분하며 부인했고 ...''

{\tiny \textcolor{gray}{
(“It was revealed that he was officially accused of manipulating a Wikipedia profile ...
and he angrily denied it, saying it was false from beginning to end.”)
}} \\

\midrule

E2E &
``위키백과 프로필이 조작되었다는 \textit{의혹이 제기된 사건}에서 ...
`사실과 전혀 다르다'며 강력히 부인했습니다.''

{\tiny \textcolor{gray}{
(“In the case where suspicions were raised about manipulation of a Wikipedia profile ...
he strongly denied it, saying it was completely untrue.”)
}} \\

\bottomrule
\end{tabular}

\vspace{-2mm}

\caption{
Qualitative analysis showing semantic divergence across MTXLS methods. Despite using the same backbone model, different methods vary in factual certainty and narrative framing, such as whether the event is described as an allegation or as an asserted fact.
}

\label{tab: inconsistency-across-methods}

\vspace{-4mm}
\end{table}
% \begin{table}[t]
% \centering
% \scriptsize
% \setlength{\tabcolsep}{3pt}

% \begin{tabular}{p{1.0cm}|p{5.4cm}}
% \toprule
% \textbf{Language} & \textbf{Generated summary excerpt} \\
% \midrule

% German &
% ``... um den Vorsprung auf den zweiten Platz zu vergrößern, während Hibs im \textbf{Scottish-Cup-Semi-Finale} auf Falkirk treffen.'' \\

% \midrule

% Hungarian &
% ``... hogy a második helyen négy ponttal vezessenek, miközben a Hibs a \textbf{skót kupa-döntőjében} játszik a Falkirk ellen.'' \\

% \midrule

% Turkish &
% ``... ikinci sırada dört puan farkla yer almak hedeflediklerini söyledi. Hibs ise \textbf{Şampiyonlar Kupası yarı finalinde} Falkirk ile karşı karşıya gelecek.'' \\

% \bottomrule
% \end{tabular}

% \vspace{-2mm}

% \caption{
% Example of cross-lingual factual inconsistency in generated summaries. German correctly refers to the Scottish Cup semifinal, while Hungarian and Turkish generate hallucinated event entities.
% }

% \label{tab: inconsistency-across-languages}

% \vspace{-4mm}
% \end{table}

\begin{table}[t]
\centering
\scriptsize
\setlength{\tabcolsep}{3pt}

\begin{tabular}{p{1.1cm}|p{5.7cm}}
\toprule
\textbf{Language} & \textbf{Generated summary excerpt} \\
\midrule

Source &
``... as his side aim to go four clear in second while Hibs face Falkirk in the
\textbf{Scottish Cup semi finals}. Stuart McCall wants his side ...'' \\

\midrule

German &
``... um den Vorsprung auf den zweiten Platz zu vergrößern,
während Hibs im \textbf{Scottish-Cup-Semi-Finale} auf Falkirk treffen.''

{\tiny \textcolor{gray}{
(“... to extend the lead over second place, while Hibs will face Falkirk in the \textbf{Scottish Cup semifinal}.”)
}} \\

\midrule

Hungarian &
``... hogy a második helyen négy ponttal vezessenek,
miközben a Hibs a \textbf{skót kupa-döntőjében} játszik a Falkirk ellen.''

{\tiny \textcolor{gray}{
(“... to lead by four points over second place, while Hibs will play Falkirk in the \textbf{Scottish Cup final}.”)
}} \\

\midrule

Turkish &
``... ikinci sırada dört puan farkla yer almak hedeflediklerini söyledi.
Hibs ise \textbf{Şampiyonlar Kupası yarı finalinde} Falkirk ile karşı karşıya gelecek.''

{\tiny \textcolor{gray}{
(“... said they aimed to stay four points ahead in second place.
Meanwhile, Hibs will face Falkirk in the \textbf{Champions Cup semifinal}.”)
}} \\

\bottomrule
\end{tabular}

\vspace{-2mm}

\caption{
Example of cross-lingual factual inconsistency in generated summaries. German correctly refers to the Scottish Cup semifinal, while Hungarian and Turkish generate hallucinated event entities.
}

\label{tab: inconsistency-across-languages}

\vspace{-4mm}
\end{table}

\section{Understanding How LLMs Perform MTXLS}
\label{section: understanding}

The preceding results reveal that LLMs still exhibit limitations in MTXLS. To better understand the performance gap between XLS and monolingual summarization, we introduce a layer-wise analysis framework for examining how LLMs perform end-to-end XLS. Specifically, we leverage the logit lens~\cite{nostalgebraist2020logitlens} to project hidden states from each layer into the vocabulary space, enabling us to trace how translation and summarization behaviors emerge throughout generation. We focus on 12 \textbf{non-Latin-script} languages spanning diverse resource levels: ar, bn, my, zh, ka, el, hi, ja, km, ko, ru, and th. This analysis allows us to characterize the layer-wise dynamics of translation and summarization during MTXLS, and to identify the stages at which errors arise. Figure~\ref{fig: layer-wise-analysis-method} provides an overview of the proposed layer-wise analysis framework.

\subsection{Layer-wise translation scoring}

We first propose a \textbf{layer-wise translation score}, $S^{\mathrm{tran}}$, to analyze when translation emerges during generation. 
Specifically, for each layer $l$ and token position $t$, we extract the hidden state $\mathbf{h}_{l,t}$ and project it through the LM head into the vocabulary space, yielding a probability distribution over the vocabulary $V$. We then obtain the highest-probability token as follows:

\begin{align}
P(w \mid \mathbf{h}_{l,t})
&=
\mathrm{Softmax}(\mathbf{W}_{\mathrm{LM}}\mathbf{h}_{l,t}) \\
w^{*}_{l,t}
&=
\arg\max_{w \in V} P(w \mid \mathbf{h}_{l,t})
\end{align}
To determine whether translation has occurred, we examine whether the predicted token is written in the target-language script. Specifically, the token-level translation score $s^{\mathrm{tran}}_{l,t}$ is set to 1 if $w^{*}_{l,t} \in U_{\mathrm{tgt}}$, where $U_{\mathrm{tgt}}$ denotes the Unicode script set of the target language. 
Otherwise, we assign the target-language confidence score produced by the Lingua language identification library~\footnote{\url{https://pypi.org/project/lingua-language-detector/}}, denoted as $\mathrm{Lingua}(w^{*}_{l,t}, \mathrm{tgt})$.
We then average the token-level scores across all $N$ generated tokens to obtain the layer-wise translation score:
\begin{align}
S^{\mathrm{tran}}(l)
&=
\frac{1}{N}
\sum_{t=1}^{N}
s^{\mathrm{tran}}_{l,t}
\end{align}

\begin{figure*}[ht]
\centering
\includegraphics[width=\textwidth]{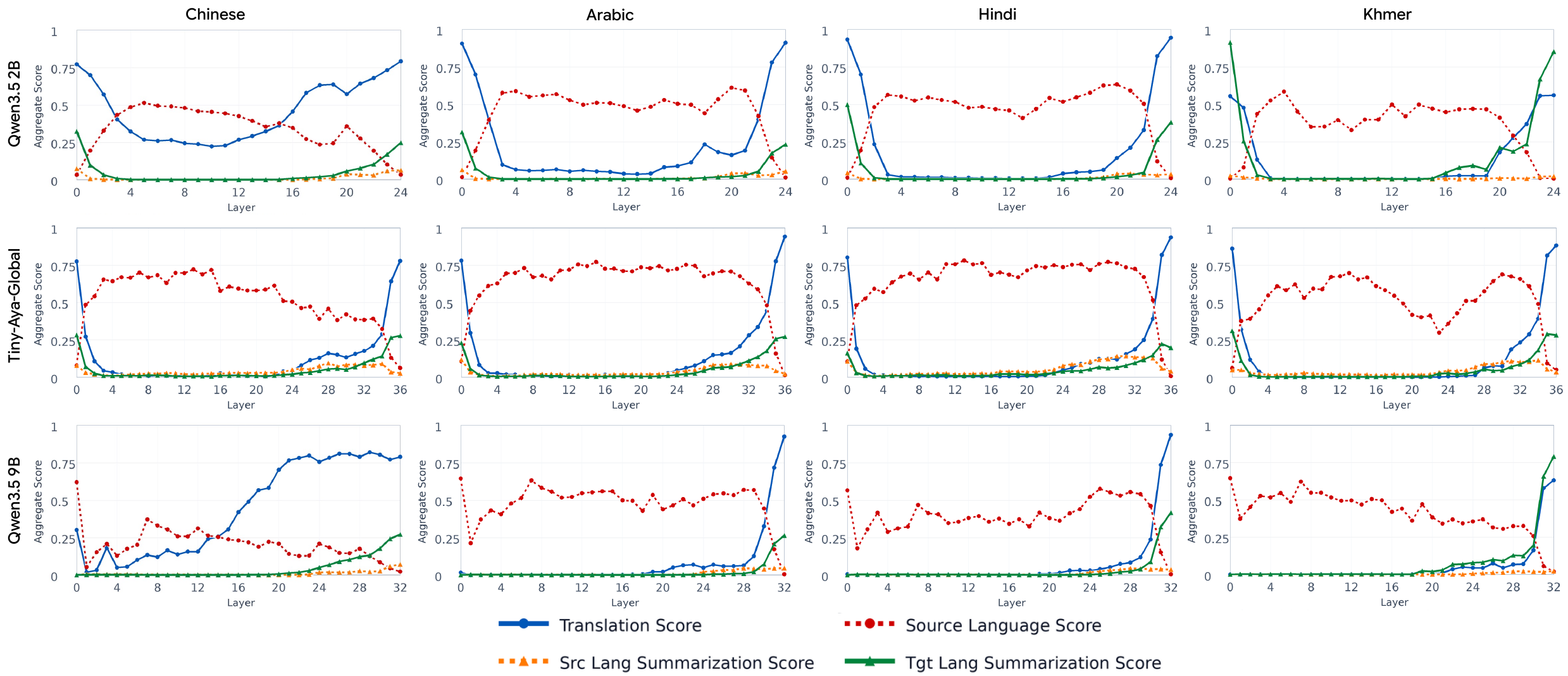}
\caption{Layer-wise MTXLS analysis results. Translation and summarization behaviors emerge around similar later layers, suggesting that MTXLS does not follow clearly separated translation and summarization stages.}
\label{fig: layer-wise-analysis-result}
\end{figure*}

\subsection{Layer-wise summarization scoring}

Similarly, we propose a \textbf{layer-wise summarization score}, $S^{\mathrm{summ}}$, to analyze when summarization emerges across layers. Following the translation analysis, for each layer $l$ and token position $t$, we use the vocabulary distribution $P(w \mid \mathbf{h}_{l,t})$ obtained from the projected hidden state. 
We then tokenize the reference summaries to construct a reference token set $R$ and restrict the vocabulary distribution to tokens in $R$. The token-level summarization score is defined as the maximum probability assigned to any token in the reference set, i.e., $\max_{w \in R} P(w \mid \mathbf{h}_{l,t})$. Finally, we compute the layer-wise summarization score by averaging across all $N$ generated tokens:
\begin{align}
S^{\mathrm{summ}}(l)
&=
\frac{1}{N}
\sum_{t=1}^{N}
\max_{w \in R} P(w \mid \mathbf{h}_{l,t})
\end{align}

Since an LLM performing XLS may either first summarize in the source language or directly generate summaries in the target language, we compute two variants of the metric using source-language and target-language references, denoted as $S^{\mathrm{summ}}_{\mathrm{src}}(l)$ and $S^{\mathrm{summ}}_{\mathrm{tgt}}(l)$, respectively.

\subsection{Layer-wise error detection}

We further investigate how errors emerge across layers through layer-wise error analysis. Specifically, we focus on two representative failure modes in summarization: \textbf{hallucination} and \textbf{omission}. Using the Recognizers-Text library~\cite{soft:recognizers-text}, we extract factual entities from the source document, reference summaries, and generated summary, denoted as $E^{\mathrm{src}}$, $E^{\mathrm{ref}}$, and $E^{\mathrm{gen}}$. We define the hallucinated entity set as $E^{\mathrm{hall}} = E^{\mathrm{gen}} \setminus (E^{\mathrm{src}} \cup E^{\mathrm{ref}})$ and the omitted entity set as $E^{\mathrm{omit}} = E^{\mathrm{ref}} \setminus E^{\mathrm{gen}}$.

For each entity $e$, we tokenize its entity string into a set of token ids $V(e)$. At each layer $l$ and decoding position $t$, we measure the visibility of error-related entities as the probability mass assigned to their constituent token ids:
\begin{align}
S^{\mathrm{err}}(l)
=
\frac{1}{|E|}
\sum_{e \in E}
\frac{1}{T}
\sum_{t=1}^{T}
\sum_{v \in V(e)}
P(v \mid \mathbf{h}_{l,t}),
\end{align}
where $E$ denotes either $E^{\mathrm{hall}}$ or $E^{\mathrm{omit}}$ depending on the error type.

\section{Analyzing Layer-wise MTXLS Behavior}

Analyzing model behavior through embedding trajectories derived from layer-wise hidden states has been explored across various domains~\cite{wang2024embedding, liu-etal-2025-tracing, wang-etal-2025-lost-multilinguality, wang-etal-2025-language-mixing, lu-etal-2025-paths, seo2026behavior}. Building on our layer-wise scoring framework introduced earlier,
% in the previous section
we analyze the model behavior during MTXLS.

\subsection{Translation emergence across layers}

Figure~\ref{fig: layer-wise-analysis-result} presents the target-language translation score alongside the source-language score by layers. Across models, most languages exhibit broadly consistent trends.
For Qwen3.5-2B, the translation score begins increasing around layer 20 before sharply rising at layer 23. Qwen3.5-9B shows a similar pattern, with translation signals emerging around layer 20 and rapidly increasing at layer 29.
Chinese exhibits a distinct behavior. Since Qwen models are predominantly trained on Chinese data, both Qwen3.5-2B and Qwen3.5-9B exhibit strong Chinese translation signals markedly earlier than other target languages, with the dominant internal language shifting from English to Chinese in the middle layers.

In contrast, Tiny-Aya-Global, which is trained more uniformly across languages, exhibits highly consistent behavior across most target languages. Translation signals gradually emerge around layer 28 and sharply increase at layer 35.
Overall, these results suggest that \textbf{each model has characteristic layers at which translation emerges, with most models exhibiting sharp translation-related transitions in later layers}, similar to patterns observed in prior work~\cite{wang-etal-2025-lost-multilinguality}.

% \begin{figure}[t]
% \centering
% \includegraphics[width=0.4\textwidth]{figs/fig-layer-wise-error-analysis_p.pdf}
% \caption{test.}

% \label{fig: fig-layer-wise-error-analysis}
% \end{figure}

\begin{figure*}[ht]
\centering
\includegraphics[width=\textwidth]{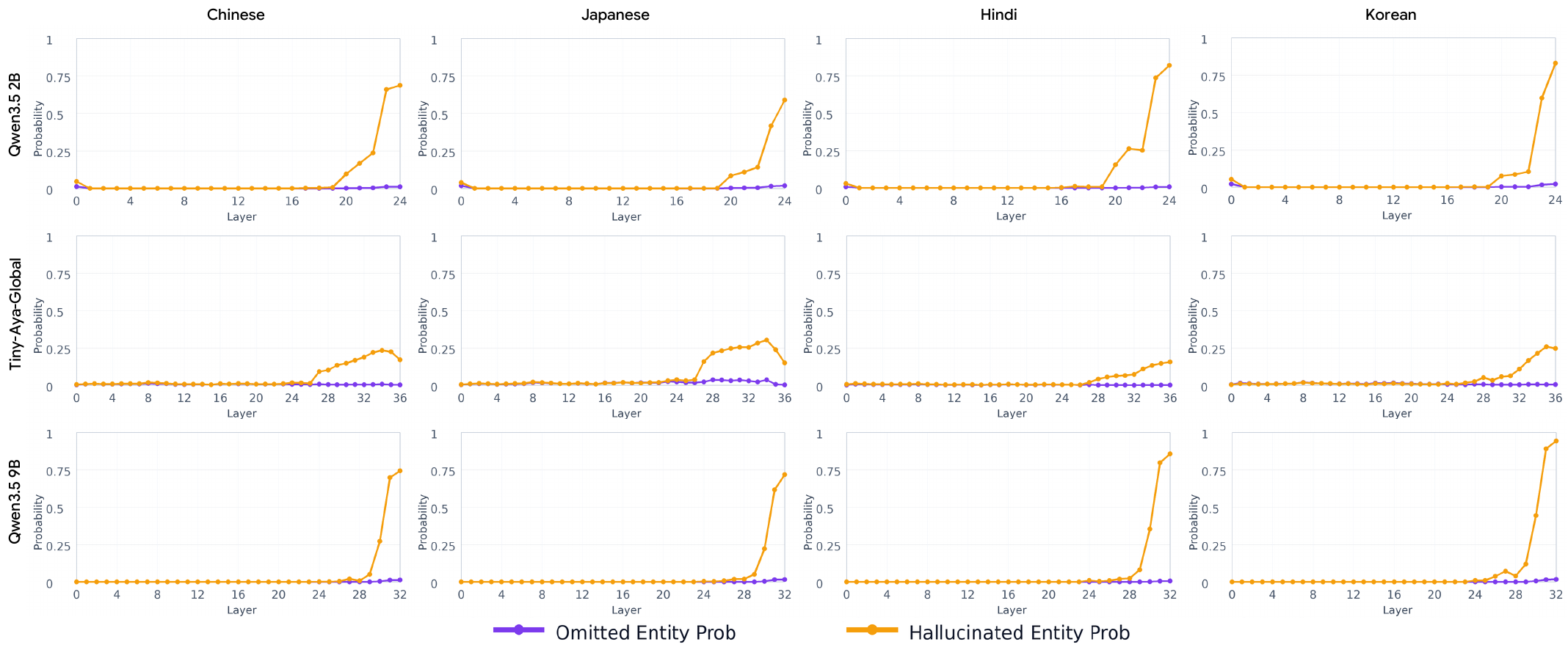}
\caption{Layer-wise error entity probability. Hallucinated entities become more probable in later layers where translation and summarization behaviors emerge, whereas omitted entities remain negligible across layers.}

\label{fig: layer-wise-error-analysis}
\end{figure*}

\subsection{Summarization emergence across layers}

Similarly, we observe consistent trends in summarization that align with the translation dynamics. Across all models, target-language summarization signals gradually emerge in the middle layers before exhibiting sharp increases in later layers. Specifically, the transition occurs around layer 23 for Qwen3.5-2B, layer 29 for Qwen3.5-9B, and layer 35 for Tiny-Aya-Global.
In contrast, source-language summarization scores exhibit markedly different behavior, remaining largely absent across layers.
These findings suggest that \textbf{translation and summarization behaviors emerge jointly within language models during cross-lingual summarization rather than as distinct stages.}

% These findings suggest that \textbf{language models might not internally separate translation and summarization into distinct stages during cross-lingual summarization}. Instead, both processes appear to emerge jointly and be performed simultaneously within the model.

\subsection{Error emergence across layers}

We analyze layer-wise token probabilities of error-related entities separately for each error type. As shown in Figure~\ref{fig: layer-wise-error-analysis}, hallucinated entities emerge at layers closely aligned with the onset of XLS behavior.
For Qwen3.5-2B, hallucinated entity probabilities begin increasing around layer 20 before sharply rising at layer 23. Qwen3.5-9B exhibits a similar trend, with gradual increases until layer 28 followed by a sharp transition at layer 29. Notably, both Qwen models generate hallucinated entities with high confidence.
Tiny-Aya-Global shows an analogous emergence pattern, with hallucinated entity probabilities increasing from layer 28 and sharply rising at layer 35, corresponding to the layers where XLS behavior emerges. However, unlike the Qwen models, Tiny-Aya-Global produces hallucinated entities with lower confidence, typically below 0.5.
In contrast, omitted entities tend to maintain near-zero probabilities across most layers and rarely exhibit a clear emergence pattern.
Overall, \textbf{hallucinated entities exhibit emergence patterns that closely align with the layers where XLS behaviors arise, whereas omitted entities maintain consistently near-zero probabilities across layers.}

\section{Leveraging English Summarization Representations}

Our analyses reveal that, during XLS, LLMs do not perform summarization and translation as independent sequential processes. Instead, both behaviors emerge jointly within specific layers, where errors also arise. In contrast, such error patterns are substantially less frequent in English summarization. 
These findings motivate using English summarization representations to guide XLS generation at the layers where errors arise.
Building on prior work~\cite{subramani-etal-2022-extracting, turner2023steering, NEURIPS2025_2e94772e} showing that activation steering can intervene on internal LLM representations to modify generation behavior, we steer MTXLS hidden representations using activations derived from English summarization.

\subsection{Activation steering for MTXLS}

Based on these findings, we propose an inference-time activation steering method that intervenes on the layers where MTXLS behaviors and errors emerge. English summarization activations encode both summarization-related and language-specific information; therefore, directly steering toward these activations may interfere with target-language generation. To preserve summarization-oriented information while reducing language-specific bias, we construct a target-language-neutralized English summarization anchor by suppressing a language-contrast direction from English summarization representations.

Given an English document $x_i$, we first generate its English summary and collect the generated-token residual-stream activations at selected hidden-state layers $l$. Let $H_{\mathrm{en},i}^{(l)}=\{h_{\mathrm{en},i,t}^{(l)}\}_{t=1}^{T_i^{\mathrm{en}}}$ denote the generated-token activations from English summarization. We also collect the corresponding baseline MTXLS activations $H_{k,i}^{(l)}$ for each target language $k$. Let $\bar{H}_{\mathrm{en},i}^{(l)}$ and $\bar{H}_{k,i}^{(l)}$ denote the mean-pooled generated-token activations from English summarization and baseline MTXLS generation, respectively. For each target language $k$ and input instance $i$, we estimate a leave-one-out language-contrast direction at layer $l$:
\begin{align}
u_{k,i}^{(l)}
&=
\frac{1}{M-1}
\sum_{j \neq i}
\left(
\bar{H}_{\mathrm{en},j}^{(l)}
-
\bar{H}_{k,j}^{(l)}
\right),
\\
v_{k,i}^{(l)}
&=
\frac{
u_{k,i}^{(l)}
}{
\left\|u_{k,i}^{(l)}\right\|_2 + \epsilon
},
\end{align}
where $M$ denotes the number of paired examples used for estimating the language-contrast direction, and $\epsilon$ is a small constant for numerical stability. We then remove this direction from each English summarization activation and mean-pool the resulting token-wise representations to obtain a target-language-neutralized anchor:
\begin{align}
\bar{A}_{\mathrm{en},k,i}^{(l)}
&=
\frac{1}{T_i^{\mathrm{en}}}
\sum_{t=1}^{T_i^{\mathrm{en}}}
\left(
h_{\mathrm{en},i,t}^{(l)}
-
\left(
h_{\mathrm{en},i,t}^{(l)\top}
v_{k,i}^{(l)}
\right)
v_{k,i}^{(l)}
\right).
\end{align}

During MTXLS generation, at each selected layer, we steer only the hidden state at the current decoding position toward this anchor:
\begin{align}
\widetilde{h}_{k,i,t}^{(l)}
&=
h_{k,i,t}^{(l)}
+
\alpha
\left(
\bar{A}_{\mathrm{en},k,i}^{(l)}
-
h_{k,i,t}^{(l)}
\right).
\end{align}
where $h_{k,i,t}^{(l)}$ denotes the MTXLS residual-stream activation at decoding step $t$, and $\alpha$ controls the interpolation strength. To preserve multilingual generation ability, we apply steering only to selected hidden-state layers $\mathcal{L}_{\mathrm{steer}} \subseteq \{1, \dots, L\}$, excluding the embedding layer. Our method operates entirely at inference time without additional training.

\subsection{Results and discussion}

\paragraph{Baselines.}

We conduct experiments using Qwen3.5-2B as the backbone model, as this model exhibited substantial performance degradation in the E2E setting (Table~\ref{tab: geval_dimensions}). 
In the layer-wise analyses presented in the previous section, Qwen3.5-2B exhibits sharp transitions in both MTXLS behavior and error-related signals at layer 23. Building on this observation, we intervene on the hidden states entering layer 23 during inference, thereby steering the representations immediately before the transition becomes pronounced.
For all languages, we apply activation steering with the steering coefficient fixed to $\alpha = 0.1$. Details of the hyperparameter search are provided in Appendix~\ref{appendix: steering_hyperparameters}.

\begin{table}[t]
\centering
\scriptsize
\setlength{\tabcolsep}{4pt}
\begin{tabular}{l|c|cccc|cc}
\toprule
\textbf{Lang.} & \textbf{Meth.} & \textbf{Coh.} & \textbf{Cons.} & \textbf{Flu.} & \textbf{Rel.} & \textbf{R-1} & \textbf{BS} \\
\midrule

\multirow{4}{*}{Arabic}
& S-T & 3.80 & 3.40 & 4.20 & 3.77 & 0.255 & 0.715 \\
& T-S & 3.73 & 3.53 & 4.12 & 3.77 & 0.190 & 0.703 \\
\cmidrule(lr){2-8}
& E2E & 3.53 & 3.23 & 4.01 & 3.55 & 0.211 & 0.703 \\
& E2E-AT & \textbf{3.62} & \textbf{3.35} & \textbf{4.16} & \textbf{3.67} & 0.210 & 0.701 \\

\midrule

\multirow{4}{*}{Korean}
& S-T & 3.54 & 3.35 & 3.47 & 3.56 & 0.316 & 0.707 \\
& T-S & 3.46 & 3.42 & 3.09 & 3.69 & 0.238 & 0.690 \\
\cmidrule(lr){2-8}
& E2E & 3.22 & 3.05 & 3.12 & 3.22 & 0.274 & 0.690 \\
& E2E-AT & \textbf{3.34} & \textbf{3.25} &\textbf{ 3.96} & \textbf{3.41} & 0.286 & 0.693 \\
\midrule

\multirow{4}{*}{Greek}
& S-T & 3.22 & 2.82 & 3.41 & 3.43 & 0.243 & 0.700 \\
& T-S & 3.04 & 2.92 & 3.48 & 3.37 & 0.204 & 0.700 \\
\cmidrule(lr){2-8}
& E2E & 2.70 & 2.67 & 3.39 & 3.00 & 0.189 & 0.689 \\
& E2E-AT & \textbf{2.86} & \textbf{2.69} & \textbf{3.63} & \textbf{3.03} & 0.195 & 0.689 \\
\midrule

\multirow{4}{*}{Bengali}
& S-T & 3.10 & 3.13 & 2.62 & 3.27 & 0.149 & 0.681 \\
& T-S & 2.91 & 3.02 & 3.71 & 3.12 & 0.135 & 0.681 \\
\cmidrule(lr){2-8}
& E2E & 2.40 & \textbf{2.63} & 1.85 & \textbf{2.68} & 0.116 & 0.664 \\
& E2E-AT & \textbf{2.45} & 2.61 & \textbf{2.48} & 2.53 & 0.118 & 0.662 \\
\midrule

\multirow{4}{*}{Georgian}
& S-T & 1.35 & 1.74 & 1.58 & 1.74 & 0.061 & 0.613 \\
& T-S & 1.23 & 1.56 & 1.51 & 1.28 & 0.078 & 0.628 \\
\cmidrule(lr){2-8}
& E2E & \textbf{1.15} & \textbf{1.34} & \textbf{1.26} & \textbf{1.18} & 0.049 & 0.590 \\
& E2E-AT & 1.06 & 1.19 & 1.16 & 1.12 & 0.038 & 0.578 \\
\bottomrule
\end{tabular}
\caption{E2E-AT denotes the activation-steered variant of E2E. \textbf{Bold} indicates the higher score within E2E.}
\label{tab: activation_steering}
\end{table}

\paragraph{Results.}

As shown in Table~\ref{tab: activation_steering}, E2E-AT consistently outperforms the standard E2E approach across most target languages. The gains are large for high-resource languages such as Arabic and Korean, where E2E-AT shows comparable performance with pipeline methods. In contrast, improvements are limited or even negative for low-resource languages; for example, activation steering further degrades performance on Georgian.

This trend is likely related to prior findings that high-resource languages tend to occupy representation spaces closer to English~\cite{wang-etal-2023-gradsim, liu2026large}. Therefore, English monolingual summarization activations transfer more effectively to high-resource languages. However, residual cross-lingual disparities may remain even after attenuating language-specific components, especially for low-resource languages. 
Future work could further investigate cross-lingual representation geometry across resource levels and develop more robust language-neutral steering methods for low-resource languages.

\section{Conclusion}

In this work, we presented a systematic study of multi-target cross-lingual text summarization (MTXLS), covering benchmark construction, empirical evaluation, layer-wise analysis, and inference-time intervention. Specifically, we introduced MEA, a new benchmark covering 24 target languages. Our experiments show that current LLMs still face substantial challenges in MTXLS, with performance lagging behind English monolingual summarization. 
To better understand this gap, we proposed a layer-wise analysis framework and found that translation and summarization behaviors emerge in later layers, while errors arise around these layers rather than in a clearly separate stage.
These findings reveal that MTXLS relies on intertwined multilingual generation behaviors and highlight where failures arise during generation. 
We further explored inference-time activation steering with summarization representations as an initial step toward using these insights for model intervention.

\section*{Limitations}

Our work has several limitations. First, evaluation for low-resource languages remains challenging due to the lack of well-established tokenizers and evaluation tools. Therefore, our primary evaluation relies on LLM-based assessment using G-Eval, while tokenization-based metrics such as ROUGE are used as supplementary references. Although we employ multiple available tokenization methods for multiple languages, which are detailed in the Appendix~\ref{appendix: evaluation_details}, ROUGE scores remain unreliable for certain low-resource languages. Moreover, LLM-based evaluation may also introduce its own biases and limitations.

Second, while activation steering consistently improves performance for high-resource languages, its effectiveness is limited for some low-resource languages. This limitation likely stem from the larger representation disparities between English and low-resource languages reported in prior work. Although our method suppresses language-specific components from English summarization representations, residual cross-lingual gaps may still persist. Future work could further investigate cross-lingual representation alignment and develop steering methods that are more robust across diverse resource settings.

\section*{Ethical Statement}

This paper focuses on multi-target cross-lingual summarization and raises no ethical concerns. All datasets used are publicly available. ChatGPT was used for grammar correction, and Codex was used as a coding assistant. All experimental design, analysis, interpretation, and research ideas were developed and validated by the authors.
\section*{Acknowledgments}

This work was supported by the IITP(Institute of Information \& Coummunications Technology Planning \& Evaluation)-ITRC(Information Technology Research Center) grant funded by the Korea government(Ministry of Science and ICT)(No. IITP-2026-RS-2024-00437866) (45\%)
; by Culture, Sports and Tourism R\&D Program through the Korea Creative Content Agency grant funded by the Ministry of Culture, Sports and Tourism in 2025 (No. RS-2025-02413038, Development of an AI-Based Korean Diagnostic System for Efficient Korean Speaking Learning by Foreigners) (45\%)
; and by Institute of Information \& communications Technology Planning \& Evaluation (IITP) grant funded by the Korea government(MSIT) (No. RS-2019-II191906, Artificial Intelligence Graduate School Program (POSTECH)) (10\%).

\bibliography{custom}

\clearpage

\appendix

\section{MEA Quality Validation}
\label{appendix: MEA_construction}

As described in Section~\ref{section: mea}, we perform quality validation on the translated summaries generated using the Google Translate API by leveraging GPT-4o-mini. The validation process evaluates four types of issues: \textit{hallucination}, \textit{language mismatch}, \textit{missing entities}, and \textit{unnatural} expressions. Each issue is further categorized into minor and major severity levels. Whenever at least one issue is detected, the corresponding summary is regenerated.
We distinguish between minor and major errors to enable a more fine-grained analysis of Google Translate API quality. minor errors refer to subtle issues that are difficult to notice even for native speakers, whereas major errors correspond to severe problems requiring correction, such as semantic distortion or ungrammatical outputs.
As shown in Table~\ref{tab: error_statistics}, the proportion of Major errors remains below 2.5\% across all languages. Notably, relatively higher error rates are observed for the low-resource languages Georgian and Khmer. The prompts used for evaluation and regeneration are provided in Figure~\ref{fig: translation_review_prompt}. The resource-level categorization of languages used throughout this work is provided in Table~\ref{tab: language_resource_level}.

\section{Evaluation Details}
\label{appendix: evaluation_details}

We primarily use G-Eval, an LLM-based evaluator known to exhibit the strongest correlation with human judgment, as our main evaluation metric, and conduct all experiments with $n=20$. As mentioned in prior work, ROUGE does not align sufficiently well with human evaluation~\cite{scialom-etal-2021-questeval, zhong-etal-2022-towards, ryu-etal-2024-multi}. This limitation becomes even more pronounced in multilingual settings, where well-established tokenizers are often unavailable for many languages. Furthermore, \texttt{multilingual ROUGE}~\footnote{\url{https://pypi.org/project/multilingual-rouge/}} supports only a limited set of languages, making it insufficient for comprehensive MTXLS evaluation. Nevertheless, for comparability with future research, we additionally report ROUGE and BERTScore results.

For multilingual ROUGE, tokenization was handled according to the target language using \texttt{multilingual-rouge}. Stemming was enabled for languages with available stemming support, including English, Arabic, Bengali, Dutch, French, German, Hindi, Hungarian, Italian, Portuguese, Russian, Spanish, and Turkish. For Chinese, Japanese, and Thai, we used the library without stemming. For Korean, we used \texttt{kiwipiepy}~\footnote{\url{https://pypi.org/project/kiwipiepy/}} for tokenization before computing ROUGE. For Greek, Persian, Polish, Ukrainian, and Vietnamese, we used NLTK's \texttt{wordpunct\_tokenize}~\footnote{\url{https://www.nltk.org}} as a tokenizer. For Khmer, we used the default tokenizer in \texttt{multilingual-rouge}.

\begin{table}[t]
\centering
\scriptsize
\setlength{\tabcolsep}{3pt}
\begin{tabular}{lrrrrrrr}
\toprule
\textbf{Lang.} & \textbf{Ex.} & \textbf{Minor} & \textbf{Major} & \textbf{Hall.} & \textbf{Lang.} & \textbf{Miss.} & \textbf{Unnat.} \\
\midrule
arabic     & 200 & 3  & 1 & 4 & 0  & 1 & 0 \\
bengali    & 200 & 6  & 5 & 2 & 11 & 3 & 10 \\
burmese    & 200 & 6  & 1 & 1 & 3  & 1 & 4 \\
chinese    & 200 & 2  & 0 & 2 & 0  & 0 & 0 \\
dutch      & 200 & 5  & 1 & 3 & 0  & 3 & 0 \\
french     & 200 & 4  & 0 & 1 & 0  & 2 & 1 \\
georgian   & 200 & 12 & 4 & 6 & 6  & 3 & 8 \\
german     & 200 & 7  & 0 & 2 & 0  & 4 & 2 \\
greek      & 200 & 11 & 1 & 5 & 3  & 4 & 6 \\
hindi      & 200 & 2  & 1 & 2 & 0  & 0 & 1 \\
hungarian  & 200 & 6  & 2 & 3 & 1  & 2 & 7 \\
italian    & 200 & 4  & 0 & 2 & 0  & 3 & 1 \\
japanese   & 200 & 4  & 0 & 3 & 0  & 1 & 0 \\
khmer      & 200 & 6  & 5 & 3 & 6  & 4 & 7 \\
korean     & 200 & 4  & 0 & 0 & 0  & 2 & 1 \\
persian    & 200 & 6  & 0 & 1 & 3  & 1 & 3 \\
polish     & 200 & 9  & 1 & 1 & 4  & 6 & 2 \\
portuguese & 200 & 4  & 0 & 2 & 0  & 2 & 0 \\
russian    & 200 & 2  & 1 & 3 & 0  & 1 & 0 \\
spanish    & 200 & 5  & 0 & 2 & 0  & 3 & 0 \\
thai       & 200 & 0  & 0 & 0 & 0  & 0 & 0 \\
turkish    & 200 & 3  & 2 & 4 & 0  & 2 & 0 \\
ukrainian  & 200 & 1  & 1 & 0 & 0  & 1 & 1 \\
vietnamese & 200 & 3  & 0 & 1 & 0  & 1 & 2 \\
\bottomrule
\end{tabular}
\caption{Error statistics across languages.}
\label{tab: error_statistics}
\end{table}

\begin{table}[t]
\centering
\small
\setlength{\tabcolsep}{4pt}
\begin{tabular}{p{1.6cm}p{5.3cm}}
\toprule
\textbf{Level} & \textbf{Languages} \\
\midrule
High 
& English (en), Chinese (zh), Spanish (es), French (fr), German (de), Japanese (ja), Russian (ru), Portuguese (pt), Korean (ko), Italian (it), Dutch (nl), Polish (pl), Turkish (tr), Vietnamese (vi), Arabic (ar), Ukrainian (uk) \\
\midrule
Mid 
& Hindi (hi), Greek (el), Thai (th), Hungarian (hu), Persian (fa), Bengali (bn) \\
\midrule
Low 
& Burmese (my), Khmer (km), Georgian (ka) \\
\bottomrule
\end{tabular}
\caption{Languages grouped by resource level}
\label{tab: language_resource_level}
\end{table}

\begin{figure}[t]
\centering
\begin{quote}

{\scriptsize\textbf{Cross-lingual Summarization}}

\begin{Verbatim}[breaklines=true, breakanywhere=true, fontsize=\scriptsize]
[System Prompt]
You are a helpful assistant. Your task is to generate a
high-quality summary of a given text from a source language
into the specified target language.

[User Prompt]
{src_lang} text:
{text}

Summarize the text above in {tgt_lang}:

(Optional structured output instruction for selected models)
Return ONLY a valid JSON object:
{"summary": "..."}
\end{Verbatim}

\vspace{0.5em}

{\scriptsize\textbf{Monolingual Summarization}}

\begin{Verbatim}[breaklines=true, breakanywhere=true, fontsize=\scriptsize]
[System Prompt]
You are a helpful assistant. Your task is to generate a
high-quality summary of a given text.

[User Prompt]
{src_lang} text:
{text}

Summarize the text above in {src_lang}:

(Optional structured output instruction for selected models)
Return ONLY a valid JSON object:
{"summary": "..."}
\end{Verbatim}

\vspace{0.5em}

{\scriptsize\textbf{Translation}}

\begin{Verbatim}[breaklines=true, breakanywhere=true, fontsize=\scriptsize]
[System Prompt]
You are a helpful assistant. Your task is to produce a
faithful, fluent translation of the given text into the
specified target language.

[User Prompt]
{src_lang} summary:
{text}

Translate the text above into {tgt_lang}:

(Optional structured output instruction for selected models)
Return ONLY a valid JSON object:
{"translation": "..."}
\end{Verbatim}

\end{quote}
\caption{Prompts used for translation, summarization, and cross-lingual summarization.}
\label{fig: used_prompt}
\end{figure}
\begin{figure*}[t]
\centering
\begin{quote}
\begin{Verbatim}[breaklines=true, breakanywhere=true, fontsize=\scriptsize]
[System Prompt]
You are a careful multilingual translation quality reviewer.
Return valid JSON only.
\end{Verbatim}
\begin{Verbatim}[breaklines=true, breakanywhere=true, fontsize=\scriptsize]
[User Prompt]
You are reviewing a translated gold summary.

Important principles:
- The existing translated summary is likely already good.
- Be conservative and avoid overcorrection.
- Only revise when there is a clear and meaningful problem.
- Do not rewrite for stylistic preference, paraphrasing, or minor wording differences.
- Preserve the original meaning, entities, numbers, and factual content whenever possible.

Your tasks:
1. Check whether the summary is predominantly written in {target_language}.
2. Check whether important entities, numbers, and facts from the English summary are preserved.
3. Check whether the translated summary is reasonably natural and fluent in {target_language}.
4. Check whether the translated summary introduces unsupported content, hallucinations, or factual distortions.

Revision policy:
- If the translation is acceptable overall, keep it unchanged.
- Only revise if there is a clear issue affecting language correctness, factual faithfulness, entity preservation, or readability.
- Prefer minimal edits instead of full rewriting.

Severity guideline:
- "minor": understandable but contains small factual, fluency, or language issues.
- "major": meaning distortion, major hallucination, severe language problems, or missing important facts/entities.

Return JSON only with this schema:
{
  "language_ok": true or false,
  "entity_coverage_ok": true or false,
  "naturalness_ok": true or false,
  "hallucination_free": true or false,
  "needs_revision": true or false,
  "severity": "none" or "minor" or "major",
  "issue_types": ["language", "missing_entity", "unnatural", "hallucination"],
  "reason": "short explanation",
  "corrected_summary": "string"
}

Rules:
- Set "needs_revision" to true only when revision is clearly necessary.
- If the summary is acceptable overall, keep it unchanged.
- "corrected_summary" must always be written in {target_language}.
- If no revision is needed, set "corrected_summary" equal to the original translated summary.
- Do not invent new facts or entities.
- Preserve names, numbers, dates, and core events from the English summary.
- Minor grammatical awkwardness alone does not require revision.
- Mixed-language output should set "language_ok" to false if a substantial portion is not in {target_language}.

English summary:
"""{english_summary}"""

Translated summary:
"""{translated_summary}"""

\end{Verbatim}
\end{quote}
\caption{Prompts used for reference-summary quality review and correction.}
\label{fig: translation_review_prompt}
\end{figure*}

\section{Implementation Details}
\label{appendix: implementation_details}

All experiments are conducted using greedy decoding to facilitate behavior analysis and ensure reproducibility, with the repetition penalty fixed to 1.0. We set the maximum generation length (\texttt{max\_new\_tokens}) to 1024 for instruction-tuned models and 4096 for RLMs. During inference, Qwen3.5-2B uses its default non-thinking mode, while Qwen3.5-9B is forced into non-thinking mode by prepending \texttt{<think>\textbackslash\text{n}\textbackslash\text{n}</think>} to the output prompt. For the RLMs, namely Qwen3.5-9B and gpt-oss-20b, we exclude reasoning tokens during layer-wise analysis and use only the hidden states corresponding to the actual generated summaries. The full prompts used for translation, summarization, and cross-lingual summarization are provided in Figure~\ref{fig: used_prompt}.

Although named entity recognition (NER) models are generally better suited for entity extraction, applying them in our setting is computationally impractical because entities must be extracted repeatedly at every layer and decoding position. We therefore use the rule-based text recognition library \texttt{recognizers-text} for entity extraction. Since the library supports only a limited set of languages, our layer-wise error analysis is restricted to the supported languages.

\section{Hyperparameters}
\label{appendix: steering_hyperparameters}

We tuned the steering coefficient $\alpha$ over $\{0.01, 0.02, \dots, 0.1\}$. While certain languages achieved better results with different coefficients, a value of 0.1 yielded consistently competitive performance across most settings. To avoid language-specific hyperparameter tuning and ensure a consistent experimental setup, we use a fixed steering coefficient of 0.1 throughout all experiments.

\section{Hardware Usage}
We used 8 NVIDIA L40S GPUs for inference and 8 NVIDIA RTX 3090 GPUs for layer-wise analysis.

\section{Full Comparison Results}
\label{appendix: full_results}

We present full per-language G-Eval, ROUGE, and BERTScore results in Tables~\ref{tab: full_qwen3.5_2b}, 
\ref{tab: full_tiny_aya_global},
\ref{tab: full_qwen3.5_9b}, and~\ref{tab: full_gpt_oss_20b}. 
As discussed in Section~\ref{section: benchmarking}, English monolingual summarization substantially outperforms MTXLS across all LLMs. We also observe clear performance disparities across language resource levels, with models generally achieving stronger results on high-resource languages while performance progressively degrades for low-resource languages.

\section{Additional Layer-wise Analysis}
\label{appendix: additional_layer_wise_analysis}

\begin{figure*}[ht]
\centering
\includegraphics[width=\textwidth]{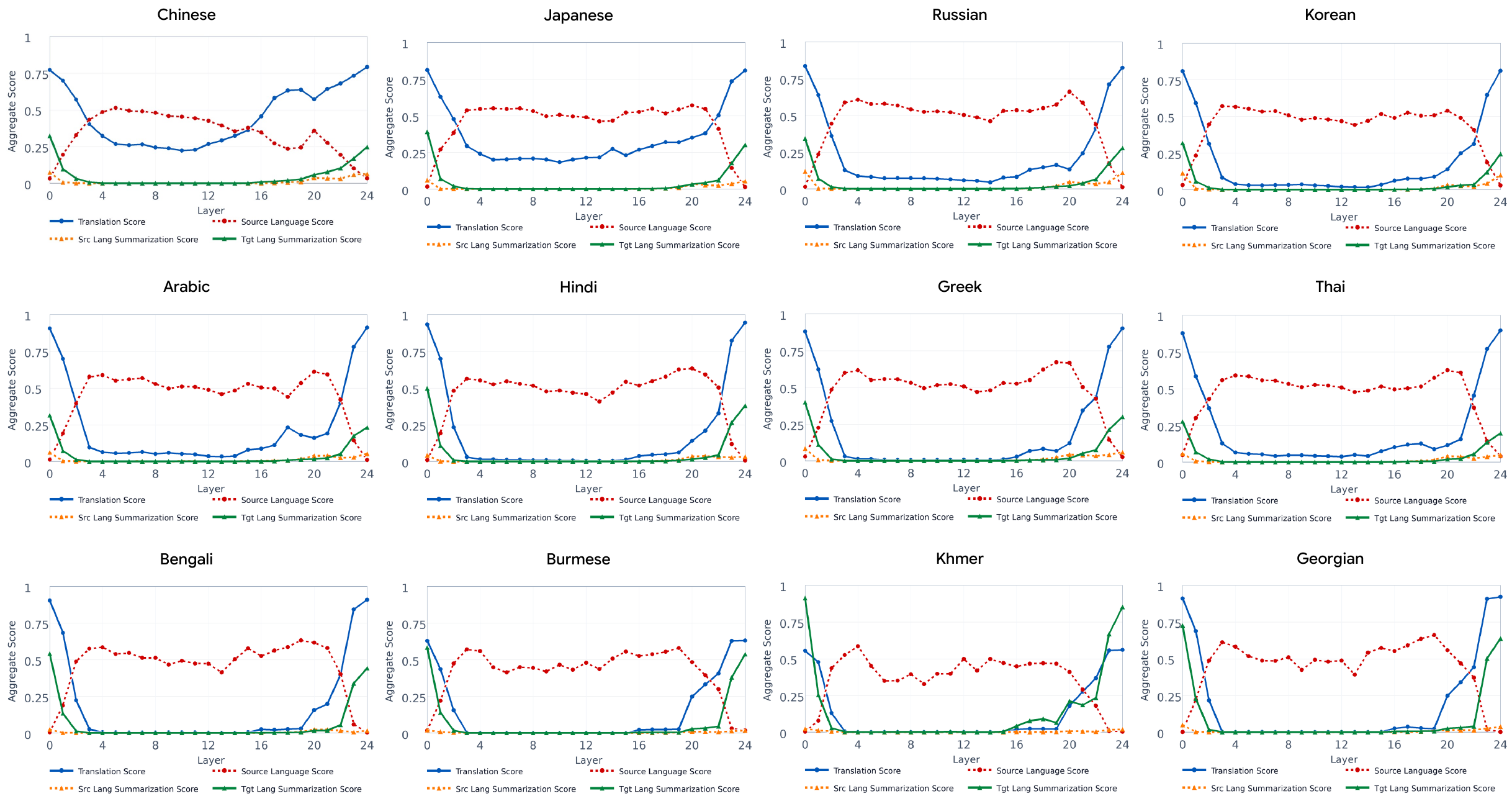}
\caption{Layer-wise MTXLS analysis results across all languages for Qwen3.5-2B.}
\label{fig: layer-wise-analysis-result-Qwen3.5-2B}
\end{figure*}

\begin{figure*}[ht]
\centering
\includegraphics[width=\textwidth]{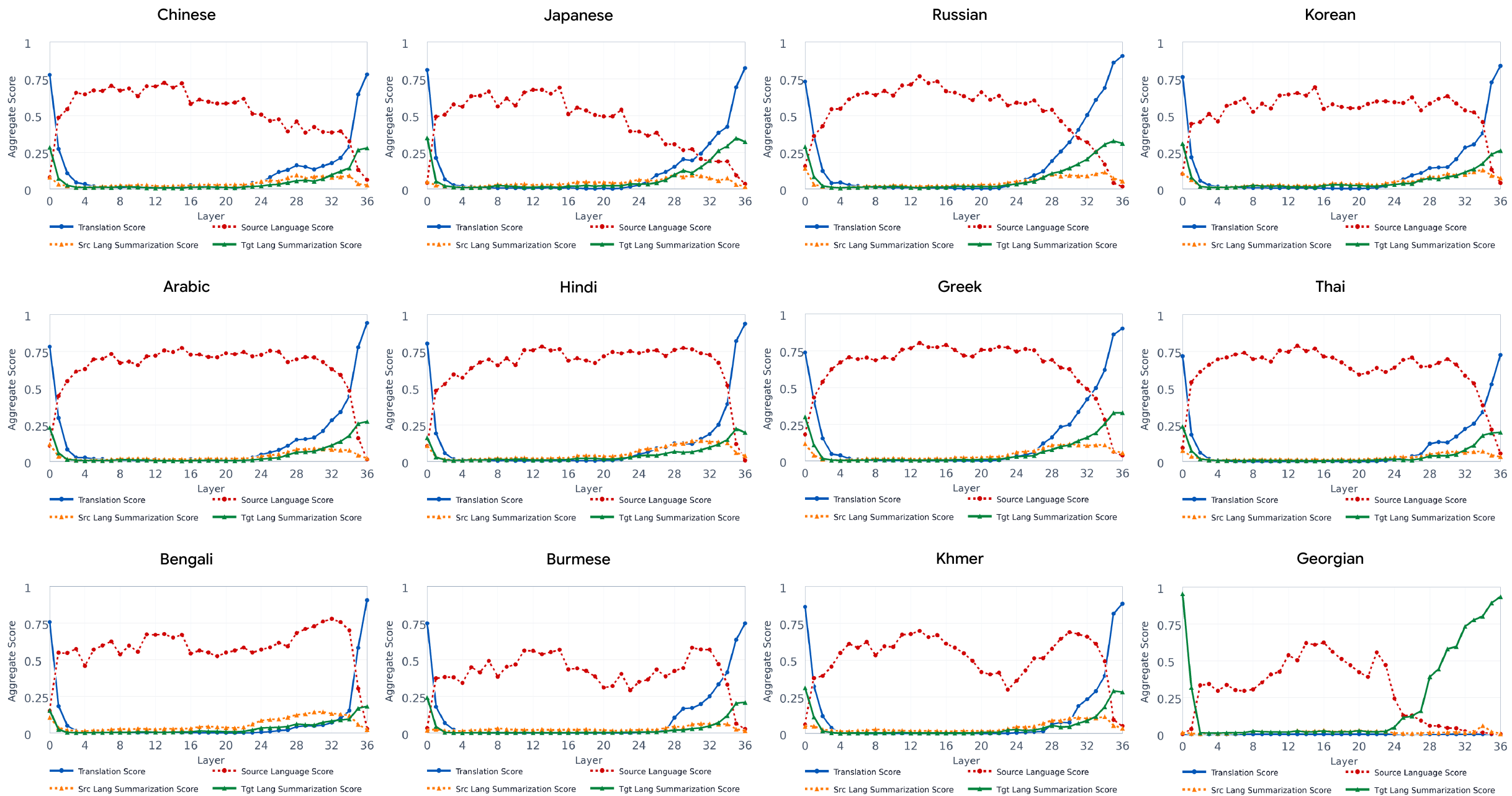}
\caption{Layer-wise MTXLS analysis results across all languages for Tiny-Aya-Global.}
\label{fig: layer-wise-analysis-result-Tiny-Aya-Global}
\end{figure*}

\begin{figure*}[ht]
\centering
\includegraphics[width=\textwidth]{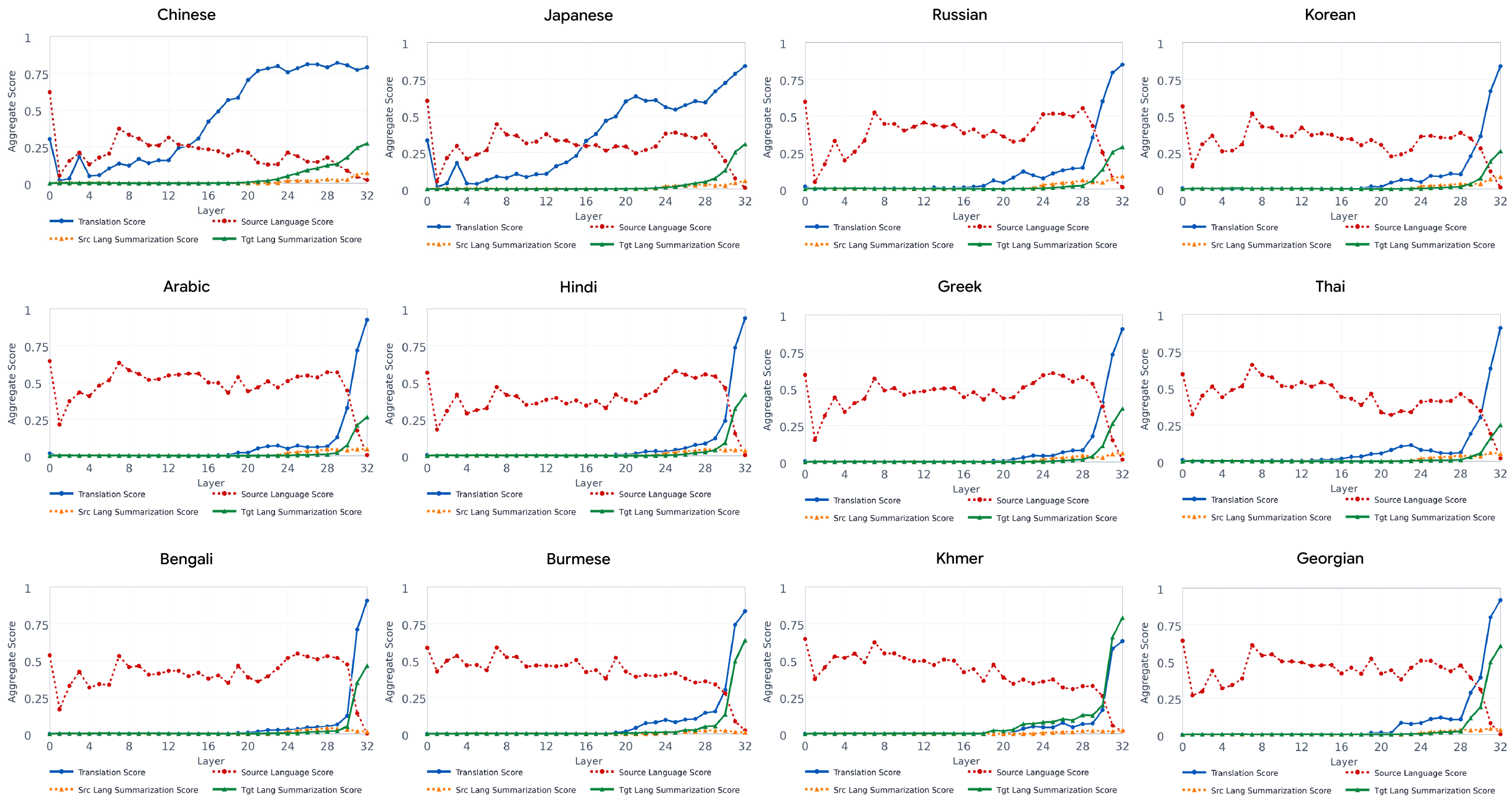}
\caption{Layer-wise MTXLS analysis results across all languages for Qwen3.5-9B.}
\label{fig: layer-wise-analysis-result-Qwen3.5-9B}
\end{figure*}

\begin{figure*}[ht]
\centering
\includegraphics[width=\textwidth]{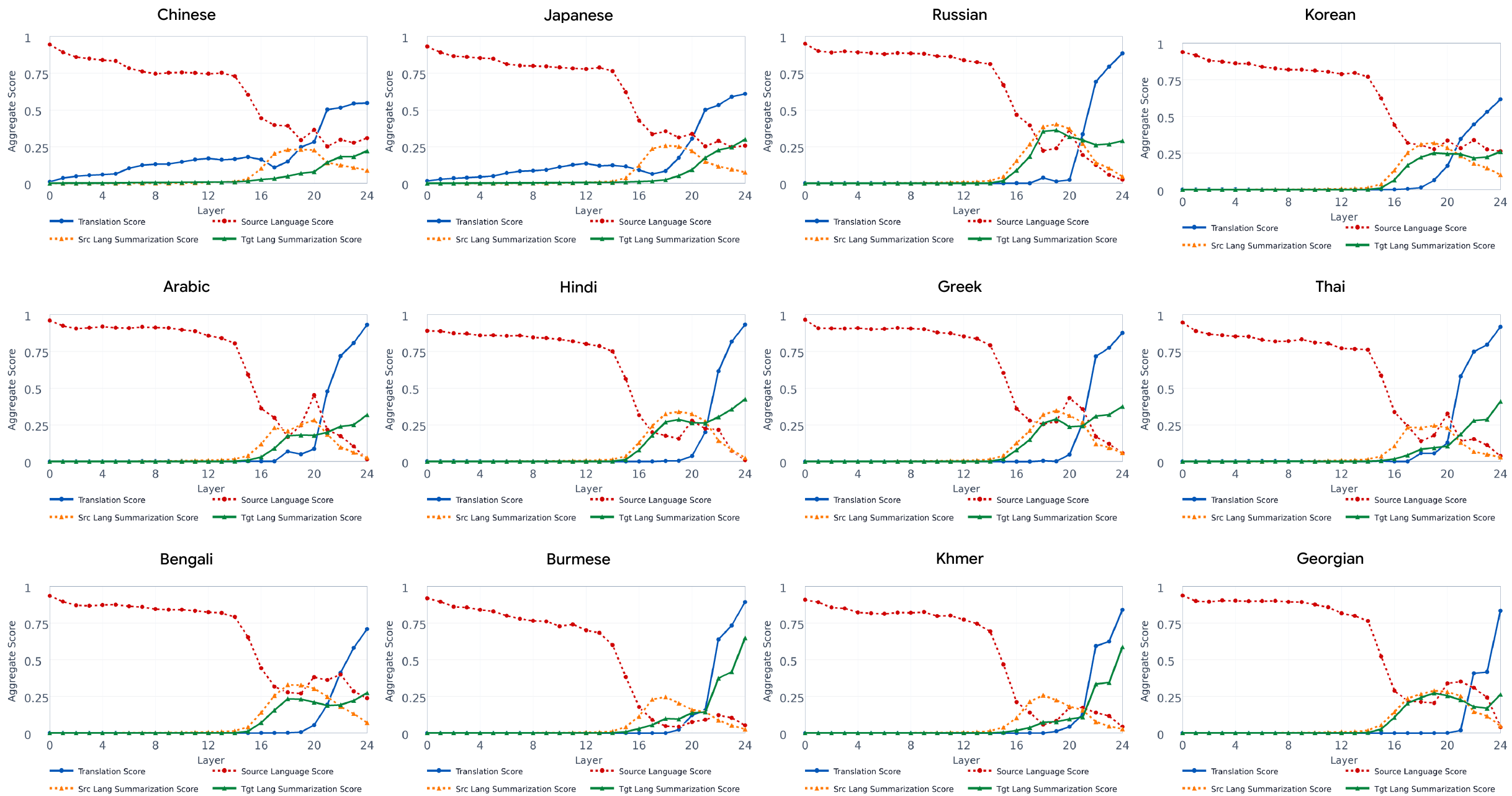}
\caption{Layer-wise MTXLS analysis results across all languages for gpt-oss-20b.}
\label{fig: layer-wise-analysis-result-gpt-oss-20b}
\end{figure*}

We provide the full layer-wise analysis results from Section~\ref{section: understanding} in Figure~\ref{fig: layer-wise-analysis-result-Qwen3.5-2B}, \ref{fig: layer-wise-analysis-result-Tiny-Aya-Global}, \ref{fig: layer-wise-analysis-result-Qwen3.5-9B}, and \ref{fig: layer-wise-analysis-result-gpt-oss-20b}. In the Qwen3.5 series, most target languages exhibit broadly consistent trends across models, with the exception of Chinese. Interestingly, in the instruction-tuned models Qwen3.5-2B and Tiny-Aya-Global, target-language signals briefly emerge in the early layers before source-language signals become dominant from the middle layers onward. In contrast, the reasoning models Qwen3.5-9B and gpt-oss-20b are dominated by source-language signals from the early layers, with little evidence of target-language signals throughout the network.

\begin{table*}[t]
\centering
\scriptsize
\scalebox{1}{
\begin{tabular}{l|c|ccccc|cc|c}
\hline
Language & Method & Coh. & Con. & Flu. & Rel. & Avg. & ROUGE-1 & ROUGE-L & BERTScore-F1 \\
\hline

English & - & 4.13 & 4.42 & 4.94 & 4.58 & 4.52 & 0.385 & 0.242 & 0.750 \\
\hline

\multirow{3}{*}{Chinese}
& S-T & 4.03 & 4.11 & 4.90 & 4.27 & 4.33 & 0.311 & 0.217 & 0.734 \\
& T-S & 4.18 & 4.34 & 4.90 & 4.43 & 4.46 & 0.260 & 0.183 & 0.723 \\
& E2E & 4.09 & 4.25 & 4.88 & 4.39 & 4.40 & 0.287 & 0.198 & 0.729 \\
\hline

\multirow{3}{*}{Spanish}
& S-T & 4.05 & 4.10 & 4.76 & 4.32 & 4.31 & 0.375 & 0.232 & 0.744 \\
& T-S & 4.14 & 4.41 & 4.71 & 4.45 & 4.43 & 0.284 & 0.184 & 0.726 \\
& E2E & 4.09 & 4.26 & 4.80 & 4.40 & 4.39 & 0.324 & 0.209 & 0.735 \\
\hline

\multirow{3}{*}{French}
& S-T & 4.03 & 3.95 & 4.49 & 4.20 & 4.17 & 0.363 & 0.218 & 0.743 \\
& T-S & 3.98 & 4.08 & 4.36 & 4.23 & 4.16 & 0.303 & 0.193 & 0.732 \\
& E2E & 4.00 & 4.00 & 4.47 & 4.20 & 4.17 & 0.309 & 0.189 & 0.733 \\
\hline

\multirow{3}{*}{German}
& S-T & 3.98 & 3.85 & 4.17 & 4.05 & 4.01 & 0.303 & 0.183 & 0.719 \\
& T-S & 3.94 & 3.94 & 4.22 & 3.98 & 4.02 & 0.224 & 0.139 & 0.698 \\
& E2E & 3.95 & 3.88 & 4.18 & 4.02 & 4.01 & 0.255 & 0.156 & 0.707 \\
\hline

\multirow{3}{*}{Japanese}
& S-T & 3.79 & 3.68 & 3.53 & 3.80 & 3.70 & 0.343 & 0.217 & 0.718 \\
& T-S & 3.65 & 3.68 & 4.12 & 3.87 & 3.83 & 0.247 & 0.165 & 0.705 \\
& E2E & 3.47 & 3.48 & 3.12 & 3.57 & 3.41 & 0.289 & 0.189 & 0.705 \\
\hline

\multirow{3}{*}{Russian}
& S-T & 3.88 & 3.68 & 4.51 & 3.99 & 4.01 & 0.281 & 0.181 & 0.713 \\
& T-S & 3.81 & 3.76 & 4.53 & 4.03 & 4.03 & 0.253 & 0.166 & 0.710 \\
& E2E & 3.81 & 3.63 & 4.45 & 3.95 & 3.96 & 0.259 & 0.169 & 0.708 \\
\hline

\multirow{3}{*}{Portuguese}
& S-T & 4.08 & 4.12 & 4.83 & 4.24 & 4.32 & 0.358 & 0.221 & 0.741 \\
& T-S & 4.11 & 4.26 & 4.79 & 4.31 & 4.37 & 0.307 & 0.197 & 0.732 \\
& E2E & 4.10 & 4.20 & 4.73 & 4.28 & 4.33 & 0.302 & 0.186 & 0.730 \\
\hline

\multirow{3}{*}{Korean}
& S-T & 3.54 & 3.35 & 3.47 & 3.56 & 3.48 & 0.316 & 0.195 & 0.707 \\
& T-S & 3.46 & 3.42 & 3.09 & 3.69 & 3.41 & 0.238 & 0.154 & 0.690 \\
& E2E & 3.22 & 3.05 & 3.12 & 3.22 & 3.15 & 0.274 & 0.174 & 0.690 \\
\hline

\multirow{3}{*}{Italian}
& S-T & 4.01 & 3.94 & 4.53 & 4.18 & 4.16 & 0.343 & 0.209 & 0.737 \\
& T-S & 3.96 & 4.02 & 4.36 & 4.14 & 4.12 & 0.290 & 0.182 & 0.726 \\
& E2E & 4.01 & 3.92 & 4.51 & 4.14 & 4.14 & 0.317 & 0.194 & 0.730 \\
\hline

\multirow{3}{*}{Dutch}
& S-T & 3.93 & 3.84 & 4.13 & 4.00 & 3.98 & 0.329 & 0.202 & 0.718 \\
& T-S & 3.82 & 3.74 & 4.08 & 3.98 & 3.90 & 0.256 & 0.165 & 0.707 \\
& E2E & 3.89 & 3.63 & 4.05 & 3.98 & 3.89 & 0.302 & 0.189 & 0.716 \\
\hline

\multirow{3}{*}{Polish}
& S-T & 3.74 & 3.36 & 4.12 & 3.88 & 3.78 & 0.229 & 0.150 & 0.696 \\
& T-S & 3.43 & 3.38 & 3.89 & 3.77 & 3.61 & 0.154 & 0.106 & 0.677 \\
& E2E & 3.38 & 3.21 & 3.74 & 3.66 & 3.49 & 0.187 & 0.127 & 0.681 \\
\hline

\multirow{3}{*}{Turkish}
& S-T & 3.16 & 3.12 & 3.46 & 3.25 & 3.25 & 0.226 & 0.151 & 0.691 \\
& T-S & 2.89 & 3.13 & 3.46 & 3.32 & 3.20 & 0.169 & 0.115 & 0.679 \\
& E2E & 2.67 & 3.01 & 3.20 & 3.11 & 3.00 & 0.187 & 0.127 & 0.679 \\
\hline

\multirow{3}{*}{Vietnamese}
& S-T & 3.99 & 3.92 & 4.79 & 4.15 & 4.21 & 0.341 & 0.203 & 0.745 \\
& T-S & 4.02 & 4.02 & 4.75 & 4.21 & 4.25 & 0.306 & 0.191 & 0.734 \\
& E2E & 4.06 & 4.08 & 4.73 & 4.22 & 4.27 & 0.289 & 0.178 & 0.730 \\
\hline

\multirow{3}{*}{Arabic}
& S-T & 3.80 & 3.40 & 4.20 & 3.77 & 3.79 & 0.255 & 0.167 & 0.715 \\
& T-S & 3.73 & 3.53 & 4.12 & 3.77 & 3.79 & 0.190 & 0.128 & 0.703 \\
& E2E & 3.53 & 3.23 & 4.01 & 3.55 & 3.58 & 0.211 & 0.140 & 0.703 \\
\hline

\multirow{3}{*}{Ukrainian}
& S-T & 3.57 & 3.19 & 4.15 & 3.74 & 3.66 & 0.198 & 0.127 & 0.700 \\
& T-S & 3.32 & 3.13 & 4.02 & 3.54 & 3.50 & 0.176 & 0.116 & 0.694 \\
& E2E & 3.20 & 2.88 & 4.05 & 3.34 & 3.37 & 0.181 & 0.117 & 0.691 \\
\hline

\multirow{3}{*}{Hindi}
& S-T & 3.46 & 3.29 & 2.18 & 3.48 & 3.10 & 0.250 & 0.146 & 0.698 \\
& T-S & 3.03 & 3.24 & 3.05 & 3.22 & 3.13 & 0.213 & 0.132 & 0.696 \\
& E2E & 3.16 & 3.22 & 2.45 & 3.23 & 3.01 & 0.238 & 0.146 & 0.695 \\
\hline

\multirow{3}{*}{Greek}
& S-T & 3.22 & 2.82 & 3.41 & 3.43 & 3.22 & 0.243 & 0.147 & 0.700 \\
& T-S & 3.04 & 2.92 & 3.48 & 3.37 & 3.20 & 0.204 & 0.132 & 0.700 \\
& E2E & 2.70 & 2.67 & 3.39 & 3.00 & 2.94 & 0.189 & 0.125 & 0.689 \\
\hline

\multirow{3}{*}{Thai}
& S-T & 3.95 & 3.85 & 4.41 & 4.13 & 4.08 & 0.325 & 0.219 & 0.731 \\
& T-S & 3.96 & 4.04 & 4.40 & 4.20 & 4.15 & 0.227 & 0.158 & 0.727 \\
& E2E & 3.93 & 3.94 & 4.40 & 4.14 & 4.10 & 0.270 & 0.184 & 0.728 \\
\hline

\multirow{3}{*}{Hungarian}
& S-T & 3.43 & 3.08 & 3.61 & 3.48 & 3.40 & 0.237 & 0.163 & 0.688 \\
& T-S & 2.78 & 2.87 & 3.13 & 3.03 & 2.95 & 0.164 & 0.117 & 0.664 \\
& E2E & 2.52 & 2.56 & 2.55 & 2.60 & 2.56 & 0.170 & 0.122 & 0.656 \\
\hline

\multirow{3}{*}{Persian}
& S-T & 3.75 & 3.54 & 3.97 & 3.74 & 3.75 & 0.268 & 0.161 & 0.710 \\
& T-S & 3.76 & 3.64 & 4.14 & 3.88 & 3.86 & 0.217 & 0.138 & 0.701 \\
& E2E & 3.74 & 3.56 & 4.14 & 3.73 & 3.79 & 0.238 & 0.149 & 0.702 \\
\hline

\multirow{3}{*}{Bengali}
& S-T & 3.10 & 3.13 & 2.62 & 3.27 & 3.03 & 0.149 & 0.100 & 0.681 \\
& T-S & 2.91 & 3.02 & 3.71 & 3.12 & 3.19 & 0.135 & 0.090 & 0.681 \\
& E2E & 2.40 & 2.63 & 1.85 & 2.68 & 2.39 & 0.116 & 0.079 & 0.664 \\
\hline

\multirow{3}{*}{Burmese}
& S-T & 1.38 & 1.61 & 1.77 & 1.70 & 1.61 & 0.061 & 0.041 & 0.577 \\
& T-S & 1.63 & 1.65 & 1.75 & 1.89 & 1.73 & 0.094 & 0.024 & 0.587 \\
& E2E & 1.08 & 1.25 & 1.63 & 1.31 & 1.32 & 0.060 & 0.043 & 0.564 \\
\hline

\multirow{3}{*}{Khmer}
& S-T & 1.19 & 1.31 & 1.48 & 1.25 & 1.31 & 0.029 & 0.027 & 0.849 \\
& T-S & 1.51 & 1.81 & 2.40 & 1.85 & 1.89 & 0.034 & 0.032 & 0.840 \\
& E2E & 1.03 & 1.19 & 1.42 & 1.08 & 1.18 & 0.024 & 0.023 & 0.860 \\
\hline

\multirow{3}{*}{Georgian}
& S-T & 1.35 & 1.74 & 1.58 & 1.74 & 1.60 & 0.061 & 0.046 & 0.613 \\
& T-S & 1.23 & 1.56 & 1.51 & 1.28 & 1.39 & 0.078 & 0.061 & 0.628 \\
& E2E & 1.15 & 1.34 & 1.26 & 1.18 & 1.23 & 0.049 & 0.039 & 0.590 \\
\hline

\end{tabular}
}
\caption{Qwen3.5-2B Multilingual evaluation results sorted by language resource level}
\label{tab: full_qwen3.5_2b}
\end{table*}
\begin{table*}[t]
\centering
\scriptsize
\scalebox{1}{
\begin{tabular}{l|c|ccccc|cc|c}
\hline
Language & Method & Coh. & Con. & Flu. & Rel. & Avg. & ROUGE-1 & ROUGE-L & BERTScore-F1 \\
\hline

English & - & 4.18 & 4.46 & 4.97 & 4.62 & 4.56 & 0.376 & 0.247 & 0.749 \\
\hline

\multirow{3}{*}{Chinese}
& S-T & 3.99 & 4.00 & 4.73 & 4.20 & 4.23 & 0.304 & 0.213 & 0.726 \\
& T-S & 4.00 & 4.06 & 4.74 & 4.16 & 4.24 & 0.314 & 0.224 & 0.732 \\
& E2E & 4.06 & 4.11 & 4.84 & 4.28 & 4.32 & 0.273 & 0.195 & 0.721 \\
\hline

\multirow{3}{*}{Spanish}
& S-T & 4.13 & 4.34 & 4.96 & 4.51 & 4.49 & 0.380 & 0.243 & 0.744 \\
& T-S & 4.11 & 4.37 & 4.83 & 4.46 & 4.44 & 0.410 & 0.260 & 0.753 \\
& E2E & 4.20 & 4.49 & 4.87 & 4.61 & 4.54 & 0.339 & 0.220 & 0.739 \\
\hline

\multirow{3}{*}{French}
& S-T & 4.15 & 4.20 & 4.74 & 4.38 & 4.37 & 0.371 & 0.229 & 0.745 \\
& T-S & 4.10 & 4.21 & 4.76 & 4.31 & 4.34 & 0.387 & 0.241 & 0.750 \\
& E2E & 4.15 & 4.30 & 4.78 & 4.41 & 4.41 & 0.347 & 0.215 & 0.741 \\
\hline

\multirow{3}{*}{German}
& S-T & 4.03 & 4.11 & 4.65 & 4.24 & 4.26 & 0.314 & 0.197 & 0.724 \\
& T-S & 4.04 & 4.13 & 4.57 & 4.19 & 4.23 & 0.319 & 0.200 & 0.726 \\
& E2E & 4.03 & 4.19 & 4.53 & 4.24 & 4.25 & 0.289 & 0.180 & 0.718 \\
\hline

\multirow{3}{*}{Japanese}
& S-T & 3.95 & 3.84 & 4.58 & 4.10 & 4.12 & 0.360 & 0.236 & 0.720 \\
& T-S & 3.89 & 3.87 & 4.00 & 4.01 & 3.94 & 0.378 & 0.247 & 0.729 \\
& E2E & 4.02 & 4.01 & 4.73 & 4.18 & 4.24 & 0.328 & 0.219 & 0.713 \\
\hline

\multirow{3}{*}{Russian}
& S-T & 4.00 & 4.05 & 4.74 & 4.25 & 4.26 & 0.300 & 0.197 & 0.718 \\
& T-S & 4.01 & 4.08 & 4.76 & 4.21 & 4.26 & 0.313 & 0.204 & 0.723 \\
& E2E & 4.02 & 4.17 & 4.72 & 4.32 & 4.31 & 0.271 & 0.181 & 0.714 \\
\hline

\multirow{3}{*}{Portuguese}
& S-T & 4.18 & 4.36 & 4.94 & 4.47 & 4.49 & 0.369 & 0.235 & 0.744 \\
& T-S & 4.21 & 4.43 & 4.86 & 4.44 & 4.48 & 0.380 & 0.242 & 0.749 \\
& E2E & 4.34 & 4.54 & 4.95 & 4.59 & 4.61 & 0.318 & 0.204 & 0.737 \\
\hline

\multirow{3}{*}{Korean}
& S-T & 3.95 & 3.85 & 4.52 & 4.06 & 4.09 & 0.346 & 0.218 & 0.724 \\
& T-S & 3.93 & 3.91 & 4.55 & 4.05 & 4.11 & 0.333 & 0.218 & 0.727 \\
& E2E & 4.05 & 4.07 & 4.75 & 4.15 & 4.25 & 0.313 & 0.203 & 0.720 \\
\hline

\multirow{3}{*}{Italian}
& S-T & 4.05 & 4.24 & 4.85 & 4.42 & 4.39 & 0.354 & 0.222 & 0.739 \\
& T-S & 4.06 & 4.30 & 4.77 & 4.33 & 4.37 & 0.361 & 0.232 & 0.742 \\
& E2E & 4.16 & 4.34 & 4.79 & 4.47 & 4.44 & 0.323 & 0.206 & 0.733 \\
\hline

\multirow{3}{*}{Dutch}
& S-T & 4.09 & 4.21 & 4.68 & 4.38 & 4.34 & 0.339 & 0.214 & 0.724 \\
& T-S & 4.03 & 4.20 & 4.64 & 4.21 & 4.27 & 0.357 & 0.226 & 0.732 \\
& E2E & 4.10 & 4.40 & 4.63 & 4.40 & 4.38 & 0.294 & 0.191 & 0.720 \\
\hline

\multirow{3}{*}{Polish}
& S-T & 4.08 & 4.01 & 4.76 & 4.36 & 4.30 & 0.250 & 0.169 & 0.708 \\
& T-S & 4.08 & 4.08 & 4.74 & 4.27 & 4.29 & 0.273 & 0.186 & 0.718 \\
& E2E & 4.16 & 4.17 & 4.71 & 4.44 & 4.37 & 0.234 & 0.161 & 0.702 \\
\hline

\multirow{3}{*}{Turkish}
& S-T & 3.98 & 3.97 & 4.36 & 4.10 & 4.10 & 0.282 & 0.191 & 0.720 \\
& T-S & 3.95 & 4.05 & 4.33 & 4.10 & 4.11 & 0.263 & 0.184 & 0.717 \\
& E2E & 3.95 & 4.02 & 4.32 & 4.07 & 4.09 & 0.248 & 0.172 & 0.713 \\
\hline

\multirow{3}{*}{Vietnamese}
& S-T & 4.08 & 4.07 & 4.88 & 4.36 & 4.35 & 0.362 & 0.229 & 0.750 \\
& T-S & 4.16 & 4.27 & 4.92 & 4.43 & 4.45 & 0.332 & 0.216 & 0.743 \\
& E2E & 4.28 & 4.46 & 4.91 & 4.59 & 4.56 & 0.288 & 0.186 & 0.733 \\
\hline

\multirow{3}{*}{Arabic}
& S-T & 4.03 & 4.04 & 4.91 & 4.32 & 4.32 & 0.290 & 0.195 & 0.724 \\
& T-S & 4.07 & 4.11 & 4.85 & 4.36 & 4.35 & 0.285 & 0.193 & 0.722 \\
& E2E & 4.11 & 4.17 & 4.85 & 4.42 & 4.39 & 0.250 & 0.168 & 0.716 \\
\hline

\multirow{3}{*}{Ukrainian}
& S-T & 4.00 & 4.04 & 4.82 & 4.19 & 4.26 & 0.229 & 0.151 & 0.710 \\
& T-S & 4.03 & 4.10 & 4.83 & 4.20 & 4.29 & 0.249 & 0.171 & 0.722 \\
& E2E & 4.11 & 4.29 & 4.86 & 4.33 & 4.40 & 0.210 & 0.139 & 0.709 \\
\hline

% -------- MID --------
\multirow{3}{*}{Hindi}
& S-T & 4.03 & 4.07 & 4.32 & 4.16 & 4.14 & 0.348 & 0.215 & 0.739 \\
& T-S & 4.13 & 4.37 & 4.64 & 4.39 & 4.38 & 0.226 & 0.149 & 0.727 \\
& E2E & 4.13 & 4.27 & 4.58 & 4.33 & 4.32 & 0.288 & 0.186 & 0.737 \\
\hline

\multirow{3}{*}{Greek}
& S-T & 4.02 & 3.95 & 4.69 & 4.18 & 4.21 & 0.296 & 0.186 & 0.726 \\
& T-S & 4.00 & 3.99 & 4.67 & 4.15 & 4.20 & 0.315 & 0.207 & 0.734 \\
& E2E & 4.08 & 4.21 & 4.79 & 4.29 & 4.34 & 0.254 & 0.166 & 0.725 \\
\hline

\multirow{3}{*}{Thai}
& S-T & 3.90 & 3.73 & 4.21 & 4.01 & 3.96 & 0.308 & 0.215 & 0.726 \\
& T-S & 3.51 & 3.62 & 3.95 & 3.78 & 3.72 & 0.217 & 0.159 & 0.705 \\
& E2E & 3.81 & 3.82 & 4.25 & 3.97 & 3.96 & 0.243 & 0.173 & 0.719 \\
\hline

\multirow{3}{*}{Hungarian}
& S-T & 4.00 & 3.99 & 4.53 & 4.12 & 4.16 & 0.275 & 0.189 & 0.707 \\
& T-S & 3.98 & 4.12 & 4.40 & 4.07 & 4.14 & 0.243 & 0.169 & 0.700 \\
& E2E & 4.03 & 4.04 & 4.59 & 4.14 & 4.20 & 0.232 & 0.165 & 0.699 \\
\hline

\multirow{3}{*}{Persian}
& S-T & 4.01 & 4.08 & 4.71 & 4.22 & 4.26 & 0.309 & 0.197 & 0.727 \\
& T-S & 4.01 & 4.05 & 4.70 & 4.15 & 4.23 & 0.327 & 0.210 & 0.735 \\
& E2E & 4.07 & 4.29 & 4.80 & 4.38 & 4.38 & 0.265 & 0.173 & 0.720 \\
\hline

\multirow{3}{*}{Bengali}
& S-T & 3.92 & 3.91 & 3.61 & 4.02 & 3.87 & 0.214 & 0.142 & 0.716 \\
& T-S & 3.87 & 4.05 & 3.84 & 4.00 & 3.94 & 0.148 & 0.101 & 0.701 \\
& E2E & 3.99 & 4.07 & 3.94 & 4.12 & 4.03 & 0.195 & 0.130 & 0.715 \\
\hline

% -------- LOW --------
\multirow{3}{*}{Burmese}
& S-T & 3.74 & 3.81 & 3.72 & 3.65 & 3.73 & 0.305 & 0.181 & 0.699 \\
& T-S & 1.98 & 2.54 & 3.35 & 2.10 & 2.49 & 0.241 & 0.153 & 0.652 \\
& E2E & 3.71 & 3.81 & 3.66 & 3.39 & 3.64 & 0.273 & 0.168 & 0.700 \\
\hline

\multirow{3}{*}{Khmer}
& S-T & 4.02 & 4.12 & 4.54 & 4.21 & 4.22 & 0.080 & 0.074 & 0.798 \\
& T-S & 1.46 & 2.01 & 1.88 & 1.59 & 1.74 & 0.051 & 0.048 & 0.815 \\
& E2E & 4.19 & 4.43 & 4.65 & 4.51 & 4.45 & 0.073 & 0.066 & 0.798 \\
\hline

\multirow{3}{*}{Georgian}
& S-T & 1.26 & 1.69 & 2.50 & 1.45 & 1.72 & 0.083 & 0.065 & 0.632 \\
& T-S & 1.28 & 1.51 & 2.83 & 1.16 & 1.69 & 0.073 & 0.058 & 0.641 \\
& E2E & 1.10 & 1.50 & 2.50 & 1.37 & 1.62 & 0.073 & 0.060 & 0.618 \\
\hline

\end{tabular}
}
\caption{Tiny-Aya-Global, Multilingual evaluation results sorted by language resource level}
\label{tab: full_tiny_aya_global}
\end{table*}
\begin{table*}[t]
\centering
\scriptsize
\scalebox{1}{
\begin{tabular}{l|c|ccccc|cc|c}
\hline
Language & Method & Coh. & Con. & Flu. & Rel. & Avg. & ROUGE-1 & ROUGE-L & BERTScore-F1 \\
\hline

English & - & 4.38 & 4.80 & 4.95 & 4.87 & 4.75 & 0.351 & 0.223 & 0.741 \\
\hline

\multirow{3}{*}{Chinese}
& S-T & 4.30 & 4.64 & 4.94 & 4.73 & 4.65 & 0.285 & 0.199 & 0.730 \\
& T-S & 4.36 & 4.77 & 4.92 & 4.79 & 4.71 & 0.278 & 0.194 & 0.726 \\
& E2E & 4.39 & 4.81 & 4.95 & 4.84 & 4.75 & 0.266 & 0.188 & 0.724 \\
\hline

\multirow{3}{*}{Spanish}
& S-T & 4.37 & 4.73 & 4.97 & 4.82 & 4.72 & 0.347 & 0.222 & 0.738 \\
& T-S & 4.34 & 4.77 & 4.95 & 4.79 & 4.71 & 0.373 & 0.231 & 0.741 \\
& E2E & 4.31 & 4.77 & 4.96 & 4.79 & 4.71 & 0.336 & 0.215 & 0.737 \\
\hline

\multirow{3}{*}{French}
& S-T & 4.37 & 4.60 & 4.90 & 4.75 & 4.65 & 0.336 & 0.210 & 0.737 \\
& T-S & 4.31 & 4.55 & 4.87 & 4.60 & 4.58 & 0.375 & 0.226 & 0.744 \\
& E2E & 4.44 & 4.66 & 4.89 & 4.76 & 4.69 & 0.341 & 0.211 & 0.738 \\
\hline

\multirow{3}{*}{German}
& S-T & 4.19 & 4.54 & 4.85 & 4.59 & 4.54 & 0.292 & 0.177 & 0.716 \\
& T-S & 4.14 & 4.52 & 4.86 & 4.61 & 4.53 & 0.309 & 0.187 & 0.721 \\
& E2E & 4.17 & 4.60 & 4.84 & 4.60 & 4.55 & 0.287 & 0.177 & 0.717 \\
\hline

\multirow{3}{*}{Japanese}
& S-T & 4.08 & 4.24 & 4.75 & 4.45 & 4.38 & 0.326 & 0.213 & 0.721 \\
& T-S & 4.13 & 4.23 & 4.83 & 4.49 & 4.42 & 0.349 & 0.224 & 0.725 \\
& E2E & 4.17 & 4.32 & 4.89 & 4.53 & 4.48 & 0.325 & 0.212 & 0.723 \\
\hline

\multirow{3}{*}{Russian}
& S-T & 4.26 & 4.56 & 4.94 & 4.71 & 4.62 & 0.280 & 0.179 & 0.714 \\
& T-S & 4.26 & 4.53 & 4.92 & 4.66 & 4.59 & 0.288 & 0.185 & 0.714 \\
& E2E & 4.26 & 4.55 & 4.91 & 4.70 & 4.60 & 0.278 & 0.178 & 0.713 \\
\hline

\multirow{3}{*}{Portuguese}
& S-T & 4.45 & 4.74 & 4.99 & 4.80 & 4.74 & 0.335 & 0.210 & 0.736 \\
& T-S & 4.46 & 4.73 & 4.95 & 4.73 & 4.72 & 0.365 & 0.223 & 0.742 \\
& E2E & 4.46 & 4.78 & 4.98 & 4.78 & 4.75 & 0.327 & 0.204 & 0.735 \\
\hline

\multirow{3}{*}{Korean}
& S-T & 4.10 & 4.19 & 4.78 & 4.32 & 4.35 & 0.327 & 0.205 & 0.722 \\
& T-S & 4.11 & 4.21 & 4.76 & 4.31 & 4.35 & 0.341 & 0.211 & 0.723 \\
& E2E & 4.14 & 4.32 & 4.73 & 4.39 & 4.39 & 0.313 & 0.196 & 0.716 \\
\hline

\multirow{3}{*}{Italian}
& S-T & 4.26 & 4.65 & 4.90 & 4.71 & 4.63 & 0.321 & 0.199 & 0.731 \\
& T-S & 4.24 & 4.67 & 4.95 & 4.74 & 4.65 & 0.341 & 0.207 & 0.735 \\
& E2E & 4.27 & 4.72 & 4.94 & 4.73 & 4.67 & 0.320 & 0.197 & 0.732 \\
\hline

\multirow{3}{*}{Dutch}
& S-T & 4.18 & 4.57 & 4.79 & 4.64 & 4.55 & 0.312 & 0.194 & 0.717 \\
& T-S & 4.15 & 4.56 & 4.80 & 4.57 & 4.52 & 0.343 & 0.210 & 0.725 \\
& E2E & 4.15 & 4.64 & 4.84 & 4.62 & 4.56 & 0.314 & 0.198 & 0.720 \\
\hline

\multirow{3}{*}{Polish}
& S-T & 4.31 & 4.61 & 4.90 & 4.73 & 4.64 & 0.232 & 0.154 & 0.704 \\
& T-S & 4.34 & 4.60 & 4.91 & 4.76 & 4.65 & 0.243 & 0.160 & 0.707 \\
& E2E & 4.37 & 4.64 & 4.88 & 4.72 & 4.65 & 0.230 & 0.151 & 0.706 \\
\hline

\multirow{3}{*}{Turkish}
& S-T & 4.08 & 4.19 & 4.62 & 4.30 & 4.30 & 0.255 & 0.170 & 0.711 \\
& T-S & 4.05 & 4.24 & 4.52 & 4.24 & 4.26 & 0.266 & 0.179 & 0.714 \\
& E2E & 4.10 & 4.28 & 4.53 & 4.39 & 4.33 & 0.251 & 0.167 & 0.710 \\
\hline

\multirow{3}{*}{Vietnamese}
& S-T & 4.22 & 4.36 & 4.96 & 4.68 & 4.56 & 0.327 & 0.199 & 0.740 \\
& T-S & 4.19 & 4.38 & 4.93 & 4.70 & 4.55 & 0.351 & 0.212 & 0.743 \\
& E2E & 4.27 & 4.52 & 4.91 & 4.78 & 4.62 & 0.318 & 0.193 & 0.736 \\
\hline

\multirow{3}{*}{Arabic}
& S-T & 4.08 & 4.37 & 4.90 & 4.58 & 4.48 & 0.265 & 0.174 & 0.718 \\
& T-S & 4.08 & 4.38 & 4.89 & 4.54 & 4.47 & 0.281 & 0.180 & 0.722 \\
& E2E & 4.10 & 4.40 & 4.89 & 4.59 & 4.49 & 0.271 & 0.173 & 0.719 \\
\hline

\multirow{3}{*}{Ukrainian}
& S-T & 4.15 & 4.51 & 4.92 & 4.55 & 4.53 & 0.218 & 0.142 & 0.710 \\
& T-S & 4.16 & 4.50 & 4.92 & 4.43 & 4.50 & 0.225 & 0.145 & 0.711 \\
& E2E & 4.17 & 4.51 & 4.88 & 4.49 & 4.51 & 0.225 & 0.144 & 0.710 \\
\hline

\multirow{3}{*}{Hindi}
& S-T & 4.07 & 4.19 & 4.69 & 4.29 & 4.31 & 0.301 & 0.184 & 0.728 \\
& T-S & 4.06 & 4.17 & 4.70 & 4.23 & 4.29 & 0.318 & 0.196 & 0.732 \\
& E2E & 4.14 & 4.28 & 4.63 & 4.33 & 4.35 & 0.286 & 0.176 & 0.732 \\
\hline

\multirow{3}{*}{Greek}
& S-T & 4.10 & 4.20 & 4.75 & 4.31 & 4.34 & 0.259 & 0.163 & 0.722 \\
& T-S & 4.11 & 4.17 & 4.77 & 4.32 & 4.34 & 0.295 & 0.184 & 0.725 \\
& E2E & 4.09 & 4.17 & 4.64 & 4.27 & 4.30 & 0.276 & 0.173 & 0.726 \\
\hline

\multirow{3}{*}{Thai}
& S-T & 4.05 & 4.21 & 4.77 & 4.57 & 4.40 & 0.316 & 0.219 & 0.741 \\
& T-S & 4.07 & 4.20 & 4.85 & 4.56 & 4.42 & 0.334 & 0.226 & 0.742 \\
& E2E & 4.06 & 4.37 & 4.87 & 4.66 & 4.49 & 0.306 & 0.211 & 0.740 \\
\hline

\multirow{3}{*}{Hungarian}
& S-T & 4.07 & 4.31 & 4.72 & 4.39 & 4.37 & 0.255 & 0.170 & 0.699 \\
& T-S & 4.09 & 4.39 & 4.76 & 4.44 & 4.42 & 0.264 & 0.177 & 0.703 \\
& E2E & 4.09 & 4.43 & 4.79 & 4.46 & 4.44 & 0.240 & 0.162 & 0.700 \\
\hline

\multirow{3}{*}{Persian}
& S-T & 4.07 & 4.31 & 4.81 & 4.50 & 4.42 & 0.273 & 0.168 & 0.718 \\
& T-S & 4.05 & 4.29 & 4.85 & 4.47 & 4.41 & 0.296 & 0.185 & 0.725 \\
& E2E & 4.07 & 4.39 & 4.87 & 4.55 & 4.47 & 0.274 & 0.172 & 0.720 \\
\hline

\multirow{3}{*}{Bengali}
& S-T & 4.08 & 4.18 & 4.60 & 4.33 & 4.29 & 0.211 & 0.137 & 0.719 \\
& T-S & 4.05 & 4.13 & 4.52 & 4.24 & 4.24 & 0.215 & 0.141 & 0.719 \\
& E2E & 4.12 & 4.26 & 4.68 & 4.41 & 4.37 & 0.201 & 0.129 & 0.718 \\
\hline

\multirow{3}{*}{Burmese}
& S-T & 3.98 & 4.05 & 4.70 & 3.98 & 4.18 & 0.293 & 0.171 & 0.710 \\
& T-S & 3.95 & 4.01 & 4.62 & 4.06 & 4.16 & 0.363 & 0.210 & 0.716 \\
& E2E & 3.83 & 4.01 & 4.53 & 4.00 & 4.09 & 0.274 & 0.163 & 0.703 \\
\hline

\multirow{3}{*}{Khmer}
& S-T & 3.96 & 4.05 & 4.49 & 4.16 & 4.16 & 0.098 & 0.086 & 0.850 \\
& T-S & 4.00 & 3.99 & 4.56 & 4.07 & 4.16 & 0.115 & 0.100 & 0.867 \\
& E2E & 3.98 & 4.15 & 4.61 & 4.19 & 4.23 & 0.097 & 0.088 & 0.822 \\
\hline

\multirow{3}{*}{Georgian}
& S-T & 3.88 & 3.96 & 4.12 & 4.16 & 4.03 & 0.159 & 0.109 & 0.706 \\
& T-S & 4.04 & 4.02 & 4.45 & 4.20 & 4.18 & 0.179 & 0.126 & 0.712 \\
& E2E & 3.98 & 4.08 & 4.10 & 4.27 & 4.11 & 0.166 & 0.115 & 0.712 \\
\hline

\end{tabular}
}
\caption{Qwen3.5-9B Multilingual evaluation results sorted by language resource level}
\label{tab: full_qwen3.5_9b}
\end{table*}
\begin{table*}[t]
\centering
\scriptsize
\scalebox{1}{
\begin{tabular}{l|c|ccccc|cc|c}
\hline
Language & Method & Coh. & Con. & Flu. & Rel. & Avg. & ROUGE-1 & ROUGE-L & BERTScore-F1 \\
\hline

English & - & 4.24 & 4.61 & 4.96 & 4.77 & 4.64 & 0.403 & 0.259 & 0.725 \\
\hline

\multirow{3}{*}{Chinese}
& S-T & 4.15 & 4.42 & 4.91 & 4.55 & 4.51 & 0.303 & 0.212 & 0.727 \\
& T-S & 3.89 & 4.06 & 4.92 & 4.12 & 4.25 & 0.278 & 0.196 & 0.702 \\
& E2E & 4.20 & 4.55 & 4.93 & 4.63 & 4.58 & 0.295 & 0.208 & 0.726 \\
\hline

\multirow{3}{*}{Spanish}
& S-T & 4.25 & 4.60 & 4.97 & 4.68 & 4.63 & 0.366 & 0.228 & 0.736 \\
& T-S & 4.00 & 4.37 & 4.96 & 4.43 & 4.44 & 0.378 & 0.236 & 0.731 \\
& E2E & 4.24 & 4.64 & 4.96 & 4.66 & 4.63 & 0.359 & 0.227 & 0.739 \\
\hline

\multirow{3}{*}{French}
& S-T & 4.30 & 4.51 & 4.82 & 4.62 & 4.56 & 0.351 & 0.214 & 0.724 \\
& T-S & 3.84 & 4.14 & 4.69 & 4.16 & 4.21 & 0.340 & 0.211 & 0.710 \\
& E2E & 4.24 & 4.48 & 4.84 & 4.57 & 4.53 & 0.355 & 0.220 & 0.730 \\
\hline

\multirow{3}{*}{German}
& S-T & 4.10 & 4.36 & 4.74 & 4.44 & 4.41 & 0.302 & 0.182 & 0.716 \\
& T-S & 3.51 & 3.72 & 4.81 & 3.64 & 3.92 & 0.268 & 0.169 & 0.689 \\
& E2E & 4.10 & 4.34 & 4.84 & 4.43 & 4.43 & 0.299 & 0.184 & 0.716 \\
\hline

\multirow{3}{*}{Japanese}
& S-T & 4.03 & 4.13 & 4.82 & 4.30 & 4.32 & 0.333 & 0.215 & 0.713 \\
& T-S & 3.68 & 3.61 & 4.93 & 3.26 & 3.87 & 0.312 & 0.194 & 0.681 \\
& E2E & 4.14 & 4.24 & 4.95 & 4.46 & 4.45 & 0.366 & 0.233 & 0.723 \\
\hline

\multirow{3}{*}{Russian}
& S-T & 4.09 & 4.24 & 4.83 & 4.42 & 4.39 & 0.281 & 0.182 & 0.705 \\
& T-S & 3.72 & 3.77 & 4.82 & 3.92 & 4.06 & 0.250 & 0.163 & 0.686 \\
& E2E & 4.11 & 4.22 & 4.78 & 4.41 & 4.38 & 0.281 & 0.184 & 0.711 \\
\hline

\multirow{3}{*}{Portuguese}
& S-T & 4.30 & 4.64 & 4.86 & 4.58 & 4.59 & 0.341 & 0.214 & 0.729 \\
& T-S & 4.02 & 4.27 & 4.98 & 4.18 & 4.36 & 0.335 & 0.211 & 0.715 \\
& E2E & 4.33 & 4.66 & 4.96 & 4.67 & 4.65 & 0.341 & 0.215 & 0.735 \\
\hline

\multirow{3}{*}{Korean}
& S-T & 4.02 & 4.01 & 4.79 & 4.11 & 4.23 & 0.378 & 0.236 & 0.718 \\
& T-S & 4.08 & 4.11 & 4.81 & 4.23 & 4.31 & 0.389 & 0.243 & 0.722 \\
& E2E & 4.06 & 4.14 & 4.93 & 4.20 & 4.33 & 0.398 & 0.250 & 0.723 \\
\hline

\multirow{3}{*}{Italian}
& S-T & 4.13 & 4.47 & 4.85 & 4.56 & 4.50 & 0.343 & 0.211 & 0.729 \\
& T-S & 3.66 & 3.99 & 4.79 & 4.04 & 4.12 & 0.325 & 0.200 & 0.710 \\
& E2E & 4.10 & 4.39 & 4.84 & 4.48 & 4.45 & 0.338 & 0.209 & 0.728 \\
\hline

\multirow{3}{*}{Dutch}
& S-T & 4.11 & 4.53 & 4.77 & 4.52 & 4.48 & 0.325 & 0.202 & 0.714 \\
& T-S & 3.46 & 3.80 & 4.71 & 3.78 & 3.94 & 0.301 & 0.190 & 0.692 \\
& E2E & 4.11 & 4.45 & 4.72 & 4.43 & 4.43 & 0.337 & 0.210 & 0.720 \\
\hline

\multirow{3}{*}{Polish}
& S-T & 4.14 & 4.32 & 4.77 & 4.52 & 4.44 & 0.235 & 0.157 & 0.700 \\
& T-S & 3.71 & 3.79 & 4.87 & 3.84 & 4.05 & 0.232 & 0.157 & 0.683 \\
& E2E & 4.19 & 4.31 & 4.82 & 4.45 & 4.44 & 0.244 & 0.166 & 0.705 \\
\hline

\multirow{3}{*}{Turkish}
& S-T & 4.06 & 4.15 & 4.56 & 4.23 & 4.25 & 0.266 & 0.178 & 0.707 \\
& T-S & 3.29 & 3.55 & 4.34 & 3.41 & 3.65 & 0.237 & 0.157 & 0.682 \\
& E2E & 4.06 & 4.13 & 4.47 & 4.23 & 4.22 & 0.265 & 0.178 & 0.705 \\
\hline

\multirow{3}{*}{Vietnamese}
& S-T & 4.04 & 4.11 & 4.81 & 4.35 & 4.33 & 0.310 & 0.193 & 0.730 \\
& T-S & 3.80 & 3.85 & 4.87 & 3.80 & 4.08 & 0.305 & 0.189 & 0.711 \\
& E2E & 4.09 & 4.15 & 4.79 & 4.36 & 4.35 & 0.340 & 0.211 & 0.738 \\
\hline

\multirow{3}{*}{Arabic}
& S-T & 3.99 & 4.19 & 4.80 & 4.33 & 4.33 & 0.247 & 0.166 & 0.701 \\
& T-S & 3.80 & 3.78 & 4.87 & 3.84 & 4.07 & 0.243 & 0.161 & 0.696 \\
& E2E & 4.03 & 4.12 & 4.82 & 4.31 & 4.32 & 0.271 & 0.179 & 0.721 \\
\hline

\multirow{3}{*}{Ukrainian}
& S-T & 4.06 & 4.17 & 4.77 & 4.25 & 4.31 & 0.215 & 0.140 & 0.704 \\
& T-S & 3.41 & 3.47 & 4.68 & 3.56 & 3.78 & 0.173 & 0.117 & 0.673 \\
& E2E & 4.04 & 4.12 & 4.70 & 4.17 & 4.26 & 0.214 & 0.140 & 0.703 \\
\hline

\multirow{3}{*}{Hindi}
& S-T & 4.07 & 4.14 & 4.76 & 4.30 & 4.32 & 0.328 & 0.199 & 0.733 \\
& T-S & 3.78 & 3.82 & 4.74 & 3.67 & 4.00 & 0.320 & 0.196 & 0.712 \\
& E2E & 4.04 & 4.12 & 4.78 & 4.22 & 4.29 & 0.332 & 0.206 & 0.735 \\
\hline

\multirow{3}{*}{Greek}
& S-T & 3.85 & 3.87 & 4.15 & 3.99 & 3.97 & 0.199 & 0.123 & 0.692 \\
& T-S & 3.48 & 3.56 & 4.26 & 3.48 & 3.69 & 0.228 & 0.145 & 0.681 \\
& E2E & 3.84 & 3.84 & 4.18 & 4.00 & 3.96 & 0.276 & 0.174 & 0.710 \\
\hline

\multirow{3}{*}{Thai}
& S-T & 3.98 & 4.11 & 4.59 & 4.34 & 4.25 & 0.257 & 0.180 & 0.717 \\
& T-S & 3.87 & 3.87 & 4.69 & 4.00 & 4.11 & 0.279 & 0.195 & 0.694 \\
& E2E & 4.04 & 4.11 & 4.70 & 4.32 & 4.29 & 0.304 & 0.210 & 0.726 \\
\hline

\multirow{3}{*}{Hungarian}
& S-T & 3.91 & 4.02 & 4.29 & 4.03 & 4.06 & 0.232 & 0.159 & 0.687 \\
& T-S & 3.28 & 3.36 & 4.44 & 3.26 & 3.59 & 0.216 & 0.153 & 0.665 \\
& E2E & 4.00 & 3.99 & 4.30 & 4.09 & 4.09 & 0.243 & 0.169 & 0.695 \\
\hline

\multirow{3}{*}{Persian}
& S-T & 3.93 & 4.05 & 4.55 & 4.14 & 4.17 & 0.267 & 0.164 & 0.711 \\
& T-S & 3.75 & 3.82 & 4.56 & 3.80 & 3.98 & 0.247 & 0.156 & 0.694 \\
& E2E & 3.99 & 4.13 & 4.65 & 4.22 & 4.25 & 0.274 & 0.172 & 0.717 \\
\hline

\multirow{3}{*}{Bengali}
& S-T & 4.05 & 4.16 & 4.64 & 4.24 & 4.27 & 0.214 & 0.139 & 0.713 \\
& T-S & 3.85 & 3.81 & 4.41 & 3.85 & 3.98 & 0.201 & 0.138 & 0.700 \\
& E2E & 4.06 & 4.14 & 4.74 & 4.26 & 4.30 & 0.229 & 0.154 & 0.724 \\
\hline

\multirow{3}{*}{Burmese}
& S-T & 2.79 & 3.03 & 2.34 & 2.85 & 2.75 & 0.145 & 0.086 & 0.647 \\
& T-S & 2.82 & 2.70 & 3.74 & 2.88 & 3.03 & 0.255 & 0.154 & 0.663 \\
& E2E & 2.28 & 2.62 & 2.03 & 2.31 & 2.31 & 0.166 & 0.099 & 0.646 \\
\hline

\multirow{3}{*}{Khmer}
& S-T & 3.89 & 3.91 & 4.24 & 3.89 & 3.98 & 0.060 & 0.056 & 0.704 \\
& T-S & 3.66 & 3.60 & 4.30 & 3.85 & 3.85 & 0.058 & 0.055 & 0.710 \\
& E2E & 3.81 & 3.80 & 4.23 & 3.77 & 3.90 & 0.066 & 0.060 & 0.732 \\
\hline

\multirow{3}{*}{Georgian}
& S-T & 3.54 & 3.58 & 3.79 & 3.80 & 3.68 & 0.096 & 0.070 & 0.659 \\
& T-S & 3.49 & 3.42 & 4.06 & 3.70 & 3.66 & 0.122 & 0.089 & 0.663 \\
& E2E & 3.35 & 3.41 & 3.55 & 3.60 & 3.48 & 0.124 & 0.090 & 0.662 \\
\hline

\end{tabular}
}
\caption{gpt-oss-20b Multilingual evaluation results sorted by language resource level}
\label{tab: full_gpt_oss_20b}
\end{table*}

\end{document}